%% file: main.tex
\documentclass{fairmeta}

\usepackage{graphicx}

\usepackage{amsmath,amssymb,amsthm}
\usepackage{natbib}
\usepackage{booktabs}
\usepackage{array}
\usepackage{tabularx}
\usepackage{algorithm}
\usepackage{algpseudocode}

\usepackage[table]{xcolor}
\usepackage{booktabs}
\usepackage{multirow}
\usepackage{graphicx}
\usepackage[T1]{fontenc}
\usepackage{newtxtext,newtxmath}

\usepackage{xspace}
\usepackage{tcolorbox}
\usepackage{xcolor}
\usepackage{amsmath}

\definecolor{posCaseGray}{HTML}{52606D}
\definecolor{posCaseOrange}{HTML}{B85C00}
\definecolor{posCaseBlue}{HTML}{176B9B}
\definecolor{posCaseGreen}{HTML}{24735C}

\newtcolorbox{posStudyBlock}[2]{%
  width=\linewidth,
  colback=#1!4!white,
  colframe=#1,
  colbacktitle=#1,
  coltitle=white,
  title={#2},
  fonttitle=\bfseries\small,
  fontupper=\small,
  boxrule=0.5pt,
  arc=1mm,
  boxsep=0pt,
  left=2.5mm,
  right=2.5mm,
  top=2mm,
  bottom=2mm,
  toptitle=1mm,
  bottomtitle=1mm,
  before skip=1.5mm,
  after skip=1.5mm,
  before upper={%
    \raggedright
    \setlength{\parindent}{0pt}%
    \setlength{\parskip}{3pt}%
  }
}

\newcommand{\name}{\textit{PoS}\xspace}

\definecolor{modelgray}{gray}{0.93}
\definecolor{posblue}{RGB}{232,242,255}
\definecolor{posdark}{RGB}{38,94,151}

\usepackage[
  colorlinks=true,
  linkcolor=black,
  citecolor=posdark,
  urlcolor=posdark
]{hyperref}

\newcommand{\modelicon}[1]{%
  \raisebox{-0.27em}{%
    \includegraphics[height=1.22em]{figures/logos/#1.pdf}%
  }%
  \hspace{0.34em}%
}

\newcommand{\BackboneHeader}[2]{%
  \rowcolor{modelgray}%
  \multicolumn{15}{c}{%
    \textbf{Backbone: }\modelicon{#1}\textbf{#2}%
  }\\[0.10ex]%
  \specialrule{0.45pt}{0pt}{0.35ex}%
}

\definecolor{posmain}{RGB}{220,234,249}
\definecolor{posablation}{RGB}{239,246,253}
\definecolor{postext}{RGB}{36,74,112}

\newlength{\posbginset}
\newcommand{\ColorFirstCell}[2]{%
  \multicolumn{1}{%
    @{}>{\columncolor{#1}[-\posbginset][\tabcolsep]}l%
  }{#2}%
}

\newcommand{\ColorMiddleCell}[2]{%
  \multicolumn{1}{%
    >{\columncolor{#1}[\tabcolsep][\tabcolsep]}c%
  }{#2}%
}

\newcommand{\ColorLastCell}[2]{%
  \multicolumn{1}{%
    >{\columncolor{#1}[\tabcolsep][-\posbginset]}c@{}%
  }{#2}%
}

\newcommand{\PoSMethod}{%
  \hspace{0.35em}%
  \textbf{\textcolor{postext}{PoS}}%
  \hspace{0.30em}%
  {\scriptsize\textcolor{postext}{\textsc{(Ours)}}}%
}

\setcitestyle{authoryear,round,aysep={,},yysep={;}}
\title{Beyond Memory: Harnessing Long-Horizon Agents with Explicit Belief States}

\author[1]{Yu Luo}
\author[1]{Jiamin Jiang}
\author[1]{Yimin Zuo}
\author[2]{Xidao Wen}
\author[1]{Rongchen Gao}
\author[1]{Yongqian Sun*}
\author[1]{Shenglin Zhang}
\author[2]{Guiyang Liu}
\author[2]{Cheng Zhang}
\author[2]{Fang Situ}
\author[2]{Qi Zhou}
\author[3]{Dan Pei}
\affiliation[1]{Nankai University}
\affiliation[2]{Alibaba Group}
\affiliation[3]{Tsinghua University}
\contribution[*]{Corresponding Author}

\date{October 2026}
\metadata[Project]{\url{https://luoyu100.github.io/projects/progression-of-states/project/}}
\metadata[Code]{\url{https://github.com/luoyu100/PoS}}

\abstract{%
Large language model (LLM) agents can now undertake increasingly complex tasks, but the way they organize interaction history into memory does not ensure a coherent understanding of the current world.
We introduce \name{}, an inference-time framework that constructs and continually maintains explicit belief states as the agent's decision context.
Each belief combines an estimate of the current world state with unresolved task requirements, making explicit what the agent still needs to learn and accomplish.
To keep this belief reliable and actionable, \name{} validates its consistency and monitors task progress to detect \emph{Belief Trapping}, where the agent continues to act without making meaningful progress toward the goal.
Recovery is then tailored to both the trapping pattern and the type of unresolved task requirement.
Experiments on four benchmarks spanning execution and diagnosis show that \name{} achieves the highest overall performance on every benchmark with all three LLM backbones.
Ablations demonstrate the importance of consistency validation and recovery, while context-scaling experiments show resilience to context growth.
Together, these results support belief construction and continual maintenance as a foundation for long-horizon context management beyond history retention and compression.
}

\begin{document}
\maketitle

\begingroup
\renewcommand{\thefootnote}{}
\footnotetext{Work done during an internship at Alibaba Group.}
\endgroup

\input{sections/introduction}
\input{sections/related_work}
\input{sections/methodology_v2}
\input{sections/experiments}

\input{sections/conclusion}

\clearpage

\bibliographystyle{plainnat}
\bibliography{references}

\clearpage

\input{sections/appendix}

\end{document}

%% file: sections/introduction.tex
\begin{figure*}[h]
    \centering
    \includegraphics[width=\textwidth]{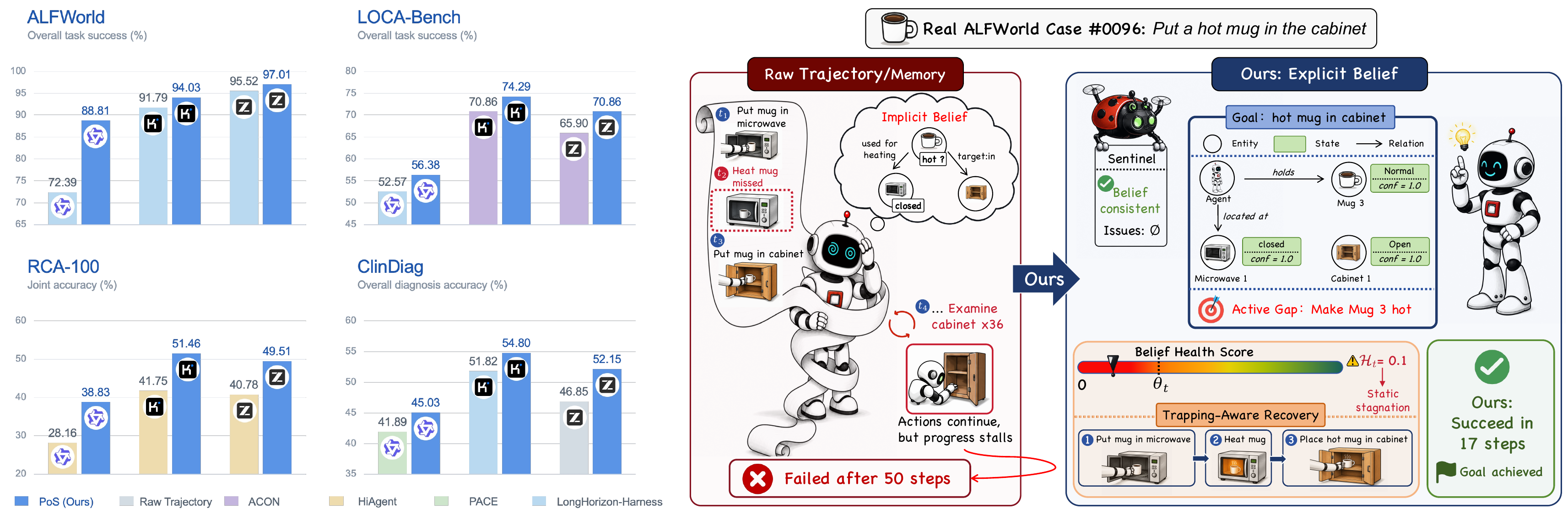}
    \caption{\textbf{Performance and motivation of \name.}
\textbf{Left:} \name outperforms the strongest baseline for each backbone across four benchmarks.
\textbf{Right:} a real ALFWorld case contrasts implicit belief reconstruction that leads to \emph{Belief Trapping} with explicit belief maintenance in \name, which enables consistency validation and recovery toward task completion.}
    \label{fig:teaser}
\end{figure*}

\section{Introduction}
\label{sec:introduction}

LLM agents are increasingly equipped with ultra-long context windows~\citep{yang2025qwen251mtechnicalreport,xu-etal-2026-128k, hu-etal-2026-every}, stronger reasoning and tool-use capabilities~\citep{Guo_2025,feng2026retool}, external memory systems~\citep{hu2025hiagent, kang2026acon, wei2026pace, xu2026mem}, and sophisticated execution harnesses~\citep{ma2026longhorizonharness,wan-etal-2026-compass}, substantially expanding the complexity and horizon of tasks they can undertake.
However, longer interaction horizons expose a deeper challenge, as actions may alter the environment and new observations may invalidate earlier inferences.
Accumulated context thus mixes evidence about past states with intermediate judgments, making it increasingly difficult to identify which facts still hold, which inferences remain supported, and what remains unresolved.
Raw trajectories preserve this evidence chronologically, while memory mechanisms select, compress, or reorganize it.
While both are valuable, access to evidence alone does not ensure an explicit and coherent estimate of the current world that can be revised and checked across steps.
Recent work observes that when task state remains implicit within a growing context, outdated or inconsistent information can persist across subsequent decisions~\citep{ma2026longhorizonharness}.
The next barrier to reliable long-horizon LLM agents is therefore not simply retaining more history, but maintaining an actionable representation of the current world.

A formal basis for this representation is the belief state \(b_t(s)=P(s_t=s \mid h_t)\), the posterior over the current latent world state \(s_t\) given interaction history \(h_t\).
In a standard partially observable Markov decision process (POMDP), this belief is a sufficient statistic for decision-making and can be recursively updated after each action and observation~\citep{kaelbling1998planning}.
Recent methods have begun to explicitly represent the belief state through goal-relative reflection, probabilistic hypotheses, structured graphs, or audited task records~\citep{kim2025reflact,tang2025trupomdp,luo2026graphstates,ma2026longhorizonharness}.
Such representations make the agent's current understanding of the world available for inspection and revision.
However, explicit belief construction is just the start, as an LLM-generated belief must satisfy two further requirements.

First, belief updates must remain internally consistent and supported by interaction evidence.
Since these updates are generated by the LLM itself, they may introduce conflicting states or unsupported inferences that undermine subsequent decisions and ultimately lead to task failure.
For instance, recording a microwave as both open and closed makes it unclear whether the robot should open it before placing a mug inside.
Continual validation is thus essential to sustain the coherence and evidential support of the evolving belief state.
Second, consistency alone does not prevent the failure mode we term \emph{Belief Trapping}, where the agent continues to act without making meaningful progress toward the goal.
For example, the agent may repeat ineffective actions, cycle through previously visited states, or gather information that does not help complete the task.
Unproductive interactions will exhaust the available budget before achieving the goal, making timely detection and recovery necessary to restore meaningful task progress.
T$^3$ detects belief deviation and truncates uninformative trajectories for reinforcement learning, but does not provide a recovery strategy within the ongoing episode~\citep{zou2026t3}.
To our knowledge, no existing framework maintains a belief state with both continual consistency validation and online diagnosis and recovery from \emph{Belief Trapping}.

We introduce \textbf{P}rogression \textbf{o}f \textbf{S}tates (\name), a unified framework that integrates explicit belief modeling, consistency validation, and trapping-aware recovery for long-horizon agent decision-making.
As illustrated in Figure~\ref{fig:overview}, \name continually updates a world state of task-relevant Entities, States, and Relations, along with epistemic and achievement gaps that specify what the agent still needs to learn or accomplish.
The agent selects actions from the current belief, while a Belief Sentinel validates updates for internal consistency and support from interaction evidence.
\name also assesses whether belief evolution supports continued task progress using three measures: gap persistence, progress stagnation, and belief recurrence.
It then performs factorized trapping diagnosis along two dimensions, agent dynamics and blocked gap type, and composes corresponding recovery constraints to recover from \emph{Belief Trapping}.

We evaluate \name using three LLM backbones on ALFWorld and LOCA-Bench for execution tasks and RCA-100 and ClinDiag for diagnosis tasks.
\name achieves the highest overall performance on all four benchmarks with each backbone, with relative gains over the strongest baseline reaching \(22.68\%\) on ALFWorld and \(37.89\%\) on RCA-100 joint accuracy.
Ablation studies demonstrate the importance of consistency validation and trapping recovery beyond explicit belief construction.
Further analyses examine trapping patterns and recovery effectiveness, robustness to context growth, and the tradeoff between token consumption and task performance.

%% file: sections/related_work.tex
\section{Related Work}
\label{sec:related-work}

We organize prior approaches to long-horizon context management according to the decision context they expose to the agent: \emph{memory as context} and \emph{belief as context}.

% TODO: Discuss memory management, state tracking, belief representations, and agent recovery.
\subsection{Memory as Context}
ReAct~\citep{yao2023react} provides a foundational framework for LLM agents, conditioning each action on the growing raw interaction trajectory.
Compression-based methods instead summarize accumulated context into compact forms: ReadAgent~\citep{lee2024readagent} constructs retrievable gist memories, ACON~\citep{kang2026acon} optimizes guidelines for history compression, and SUPO~\citep{lu-etal-2026-beyond} jointly learns periodic summarization and tool-use policies.
Other methods impose structure on the retained context: HiAgent~\citep{hu2025hiagent} organizes working memory around subgoals, COMPASS~\citep{wan-etal-2026-compass} maintains concise progress briefs across reasoning stages, and PACE~\citep{wei2026pace} adjusts historical granularity according to its predicted relevance to the next action.
Other approaches make context management itself an agent-controlled operation: MemGPT manages information across in-context and external memory through an operating-system-inspired hierarchy~\citep{packer2023memgpt}. CaT~\citep{liu-etal-2026-context} exposes context compression as a callable tool over a structured workspace, AgentFold~\citep{ye2026agentfold} performs proactive multi-scale folding of historical trajectories, and Sculptor~\citep{li2026sculptor} provides explicit tools for actively curating the working context.
Although these methods improve evidence accessibility and context efficiency, the resulting context remains a transformed record of historical evidence rather than a consistent estimate of the current world.

\subsection{Belief as Context}
A parallel line of work studies how LLMs represent the current state of an evolving world~\citep{gurnee2024space}.
Mechanistic studies probe the world-state abstractions encoded by LLMs~\citep{li2024worldrepresentations} and show that state tracking of reasoning is itself a learned computation whose underlying mechanism affects systematic generalization~\citep{li2025statetracking}.
Agent frameworks further externalize parts of this computation~\citep{tang2024worldcoder}.
QuBE~\citep{kim2024qube} constructs a task-relevant belief through question answering; StateAct~\citep{rozanov2025stateact} maintains a chain of states throughout task execution; and ReflAct~\citep{kim2025reflact} grounds action selection in reflection over the current state relative to the goal.
More structured approaches maintain particle beliefs over latent states and goals through hypothesis trees and Bayesian filtering~\citep{tang2025trupomdp}, recursively update natural-language belief summaries~\citep{lidayan2026abbel}, or combine causal graphs with state machines as explicit belief~\citep{luo2026graphstates}.
LongHorizon-Harness~\citep{ma2026longhorizonharness} externalizes task state as audited records of requirements, artifacts, and facts, preserving verified progress through a Manage-Execute-Audit loop.
However, it relies on independently verifiable intermediate results and is therefore less suitable when meaningful progress consists of uncertain inferences rather than observable outcomes.
Beyond belief construction, recent work has also begun to address failure modes related to \emph{Belief Trapping}.
T$^3$~\citep{zou2026t3} detects excessive belief deviation and truncates uninformative trajectories during reinforcement learning, while CGDP~\citep{kausik2026cgdp} uses an exhaustion gate to terminate unproductive context gathering.
These approaches address stalled progress through truncation or termination, without attempting to restore progress during inference.
In contrast, \name maintains and validates a task-conditioned belief and uses factorized recovery to redirect subsequent actions.

%% file: sections/methodology_v2.tex
\section{Methodology}
\label{sec:methodology}

% Overview 
\name comprises three components: Belief Modeling constructs a task-conditioned belief (Section~\ref{sec:belief model}), Belief-Guided Interaction selects actions and validates belief updates (Section~\ref{sec:belief-guided-Interaction}), and Trapping-Aware Recovery detects and recovers from \emph{Belief Trapping} (Section~\ref{sec:trapping-aware-recovery}).

\begin{figure*}[t]
    \centering
    \includegraphics[width=\textwidth]{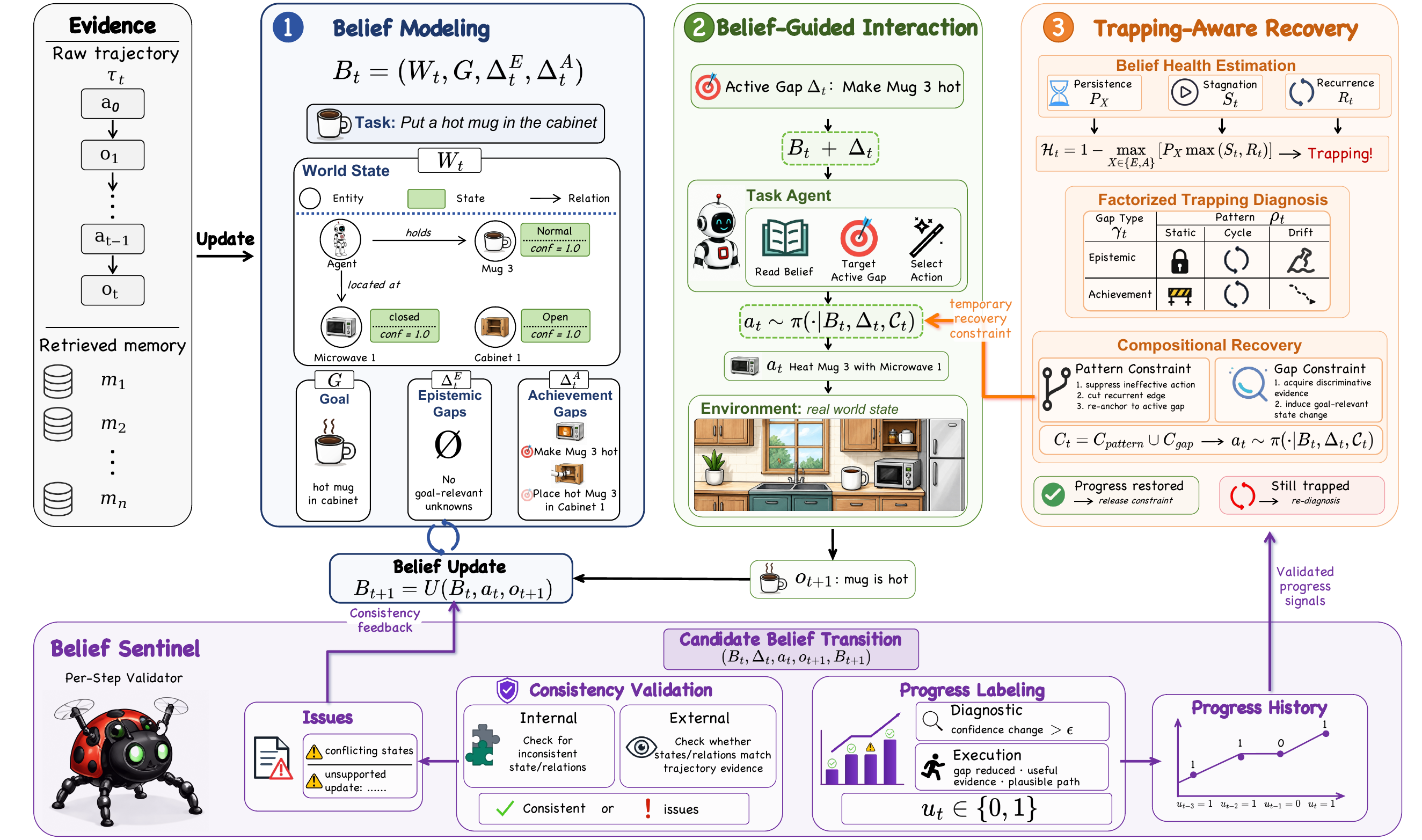}
    \caption{\textbf{Overview of \name.}
Belief Modeling constructs and incrementally updates a task-conditioned Belief from interaction evidence (Section~\ref{sec:belief model}).
Belief-Guided Interaction conditions each action on the current Belief and an Active Gap, while the Belief Sentinel validates consistency and records task-relevant progress (Section~\ref{sec:belief-guided-Interaction}).
Trapping-Aware Recovery estimates Belief health, diagnoses heterogeneous trapping conditions, and composes recovery constraints to restore progress (Section~\ref{sec:trapping-aware-recovery}).}
    \label{fig:overview}
\end{figure*}

% TODO：这里需要展示一个清晰的Problem Formulation 和 RoadMAP，这两者可以串联到一起讲；
% 如图~\ref{fig:overview} 所示，在 agent 执行长程任务的过程中，PoS 不直接将不断增长的原始交互轨迹\(\tau_t\) 或由其导出的 memory 作为 agent 的决策上下文，而是持续维护一个以任务为中心的Belief State \(B_t\)，以表征 agent 对当前世界的可行动理解。具体而言，PoS 包含三个关键部分：首先，Belief Modeling 构建以任务为中心的 Belief State，作为对当前真实世界状态的近似估计（Section~3.1）。随后，agent 在 Belief 的引导下选择下一步动作，同时引入异步运行的Sentinel 对 Belief 的一致性进行校验（Section~3.2）。最后，PoS 持续感知当前 Belief 是否陷入 trapping；一旦检测到 trapping，便及时执行 recovery，以维持 agent 的有效任务推进（Section~3.3）。

\subsection{Belief Modeling}
\label{sec:belief model}

The interaction between a long-horizon agent and its environment can be formulated as a POMDP.
At step \(t\), the environment has a latent world state \(s_t\) that is not directly observable.
Given the interaction history \(h_t\), the belief state \(b_t(s)\) represents the posterior distribution over \(s_t\) and provides the basis for selecting an action \(a_t\).
Following this formulation, \name maintains a structured, task-conditioned belief \(B_t\) as the agent's explicit decision context:
\begin{equation}
    B_t = \left(W_t, G, \Delta_t^E, \Delta_t^A\right),
    \label{eq:belief-state}
\end{equation}
where \(W_t\) is a structured estimate of the task-relevant world state, \(G\) specifies the task objective, and \(\Delta_t^E\) and \(\Delta_t^A\) capture unresolved epistemic and achievement gaps, respectively.
% Together, \(G\), \(\Delta_t^E\), and \(\Delta_t^A\) make \(W_t\) actionable by identifying the information and state changes required to advance the task.

\paragraph{World state (\(W_t\)).}
Prior work often represents an environment through entities and their relations~\citep{anokhin2025arigraph,gu2024conceptgraphs}.
For long-horizon decision making, however, entity identity and connectivity alone are insufficient: an entity's condition may change over time, directly affecting which actions are feasible and whether the task goal is satisfied.
\name therefore represents \(W_t\) with an \textit{Entity--State--Relation} structure that applies across domains and makes such changes explicit.
Within this structure, an \textit{Entity} denotes an object or agent involved in the task, while a \textit{State} describes the agent's current understanding of that Entity in natural language.
A directed \textit{Relation} captures how Entities and States are connected, including relations between two Entities, two States, or an Entity and a State.
Each State and Relation is annotated with its provenance and confidence, together with supporting evidence when available.
The natural-language descriptions in \(W_t\) provide the flexibility to express domain-specific semantics, while the explicit structure supports consistent belief updates and validation.
Rather than attempting to model the environment exhaustively, \(W_t\) retains only records that can affect action feasibility, outcome assessment, or satisfaction of \(G\).

\paragraph{Goal-conditioned components (\(G\), \(\Delta_t^E\), and \(\Delta_t^A\)).}
Although \(W_t\) captures the task-relevant world state, it does not by itself indicate what remains unresolved for completing the task.
\name evaluates \(W_t\) against the task objective and completion criteria specified by \(G\) and represents the unresolved requirements as two complementary gap sets.
Epistemic gaps \(\Delta_t^E\) capture information that remains unknown or insufficiently supported, whereas achievement gaps \(\Delta_t^A\) capture discrepancies between the current belief and the desired goal state that must be resolved through action.
This distinction separates what the agent still needs to learn from what it still needs to accomplish.
Together, \(W_t\), \(G\), \(\Delta_t^E\), and \(\Delta_t^A\) allow \(B_t\) to represent both the agent's current understanding of the world and the remaining requirements for goal completion, making it an actionable decision state.

\subsection{Belief-Guided Interaction}
\label{sec:belief-guided-Interaction}

Although \(B_t\) represents both the current world and the unresolved task requirements, it may contain multiple epistemic and achievement gaps.
Giving equal priority to all gaps in \(\Delta_t^E \cup \Delta_t^A\) leaves the agent without a clear focus and may cause it to shift between gaps without making sustained progress on any of them.
\name therefore retains \(B_t\) as the complete decision state while selecting an active gap \(\Delta_t \in \Delta_t^E \cup \Delta_t^A\) that are the most valuable to address given the current world state and task objective.
The active gap establishes the current focus without discarding the broader world understanding and task context represented by \(B_t\).
We then formulate action selection as:
\begin{equation}
    a_t \sim \pi\left(\cdot \mid B_t, \Delta_t, \mathcal{C}_t\right).
    \label{eq:belief-guided-action}
\end{equation}
Here, \(\mathcal{C}_t\) denotes the recovery constraint introduced in Section~\ref{sec:trapping-aware-recovery}.
It is empty during normal interaction and is applied only when \emph{Belief Trapping} is detected, as an additional constraint on action selection.
After executing \(a_t\), the task agent receives the resulting observation \(o_{t+1}\) from the environment.

The new observation \(o_{t+1}\) provides evidence to update the belief.
The task agent incorporates this evidence into \(B_t\) to form an unverified candidate \(\widetilde{B}_{t+1}\), which cannot yet be committed as \(B_{t+1}\).
Because \(\widetilde{B}_{t+1}\) is produced by the task agent itself, it may inherit the agent's erroneous assumptions, reasoning errors, or hallucinations.
Such errors can produce two forms of inconsistency: internal and external.
Internal inconsistency arises when \(\widetilde{B}_{t+1}\) contains mutually conflicting content, such as incompatible States assigned to the same Entity.
External inconsistency arises when \(\widetilde{B}_{t+1}\) contradicts the latest observation or other available interaction evidence.
\name therefore introduces the \emph{Belief Sentinel} to validate \(\widetilde{B}_{t+1}\).
It audits newly added or modified Entities, States, Relations, and Gaps, and returns potential consistency issues to the task agent.
The task agent evaluates this feedback against the available interaction evidence and revises \(\widetilde{B}_{t+1}\) before committing the resulting belief as \(B_{t+1}\).
The complete belief update is expressed as
\begin{equation}
    \widetilde{B}_{t+1}
    =
    \mathcal{U}\left(B_t,a_t,o_{t+1}\right)
    \;\xrightarrow{\;\text{\emph{Belief Sentinel}}\;}\;
    B_{t+1},
    \label{eq:validated-belief-update}
\end{equation}
where \(\mathcal{U}\) denotes the candidate belief update performed by the task agent, and the arrow represents Belief Sentinel validation followed by evidence-based revision by the task agent.
The resulting \(B_{t+1}\) serves as the decision state for the next step and the basis for belief health estimation in Section~\ref{sec:trapping-aware-recovery}, helping reduce the risk that erroneous belief updates are mistaken for actual interaction dynamics.

\subsection{Trapping-Aware Recovery}
\label{sec:trapping-aware-recovery}     

To assess task progress, \name tracks changes in the world state and unresolved requirements across successive validated belief transitions.
When \emph{Belief Trapping} is detected, it performs factorized diagnosis of the trapping dynamics and blocked gap type, then composes corresponding recovery constraints to guide subsequent action selection.

\paragraph{Belief health estimation (\(\mathcal{H}_t\)).}
\name estimates belief health at two temporal scales.
At the step level, each validated transition
\(B_{t-1}\rightarrow B_t\) is assigned a progress label
\(u_t\in\{0,1\}\).
For diagnostic tasks, let \(c_t(x)\) denote the confidence distribution over inferred States and Relations.
We measure progress by aggregating absolute confidence change between consecutive beliefs:
\begin{equation}
d_t^{\mathrm{diag}}
=
\frac{1}{2}
\sum_x
\left|c_t(x)-c_{t-1}(x)\right|,
\qquad
u_t
=
\mathbb{I}
\left[d_t^{\mathrm{diag}}>\epsilon\right],
\label{eq:diagnostic-progress}
\end{equation}
where \(\epsilon\) is a predefined threshold that filters out minor confidence fluctuations.
For execution tasks, the \emph{Belief Sentinel} assigns \(u_t=1\) when the transition reduces the active gap \(\Delta_t\), acquires information needed to resolve it, or advances a plausible path toward the goal; otherwise, \(u_t=0\).

However, a single transition is insufficient to determine whether the agent has become trapped.
\name therefore aggregates belief dynamics over the \(K\) most recent validated transitions,
\(\mathcal{W}_t=\{t-K+1,\ldots,t\}\).
Within this window, \name measures three complementary signals.
\textit{(1) Gap Persistence} \(P_X\), where \(X\in\{E,A\}\), measures the fraction of initially unresolved gaps of type \(X\) that remain unresolved throughout the window.
\textit{(2) Progress Stagnation} \(S_t\) measures the fraction of transitions with no recorded task progress.
\textit{(3) Belief Recurrence} \(R_t\) measures how frequently the world state relevant to the active gap repeats, taking the highest recurrence rate across candidate temporal lags.
Let \(W_\tau^\Delta\) denote the projection of \(W_\tau\) onto the current active gap \(\Delta_t\), and let \(d_J\) denote the Jaccard distance.
We measure world state change using \(d_W\), the average Jaccard distance between the corresponding Entity, State, and Relation sets.
\begingroup
\setlength{\jot}{2pt}
\setlength{\abovedisplayskip}{6pt}
\setlength{\belowdisplayskip}{6pt}
\begin{subequations}
\label{eq:belief-health}
\begin{gather}
P_X
=
\frac{
\left|
\bigcap_{\tau=t-K+1}^{t}\Delta_\tau^X
\right|
}{
\left|\Delta_{t-K+1}^X\right|
},
\qquad
S_t
=
1-\frac{1}{K}
\sum_{\tau=t-K+1}^{t}u_\tau,
\label{eq:persistence-stagnation}
\\
d_W\!\left(W_i^\Delta,W_j^\Delta\right)
=
\frac{1}{3}
\sum_{\mathcal{Y}}
d_J\!\left(\mathcal{Y}_i^\Delta,\mathcal{Y}_j^\Delta\right),
\quad
R_t
=
\max_{1\leq\ell\leq L}
\frac{1}{K-\ell}
\sum_{\tau=t-K+1+\ell}^{t}
\mathbb{I}
\left[
d_W\!\left(W_\tau^\Delta,W_{\tau-\ell}^\Delta\right)
\leq\epsilon_W
\right],
\label{eq:belief-recurrence}
\\
\mathcal{H}_t
=
1-
\max_{X\in\{E,A\}}
\left[
P_X \max\left(S_t,R_t\right)
\right].
\label{eq:belief-health-score}
\end{gather}
\end{subequations}
\endgroup
We set \(P_X=0\) when \(\Delta_{t-K+1}^X=\varnothing\).
Here, \(\mathcal{Y}\in\{\mathcal{E},\mathcal{S},\mathcal{R}\}\) indexes the aligned and semantically matched Entity, State, and Relation sets, while \(L\) and \(\epsilon_W\) denote the maximum lag and recurrence threshold, respectively.
The resulting health score captures whether a persistent unresolved gap is accompanied by progress stagnation or belief recurrence. (Detailed explanations in Appendix~\ref{app:trapping-rationale})
Accordingly, \name detects \emph{Belief Trapping} when \(\mathcal{H}_t\leq\theta_H\).

\paragraph{Factorized trapping diagnosis.}
Once \emph{Belief Trapping} is detected, \name examines validated belief transitions within the same window \(\mathcal{W}_t\) to diagnose it along two complementary dimensions: the agent dynamics pattern and the blocked gap type.
First, it classifies the agent dynamics as:
\begin{equation}
\rho_t =
\begin{cases}
\mathrm{Static},
&
d_W\!\left(W_t,W_{t-1}\right)\leq\epsilon_W,
\\
\mathrm{Cycle},
&
R_t \geq \theta_R \;\land\; k_t^*>1,
\\
\mathrm{Drift},
&
d_W\!\left(W_t^\Delta,W_{t-1}^\Delta\right)\leq\epsilon_W,
\end{cases}
\label{eq:agent-dynamics-pattern}
\end{equation}
where the conditions are evaluated in order, \(k_t^*\) denotes the dominant recurrence lag underlying \(R_t\), and \(\theta_R\) is the recurrence threshold for identifying periodic behavior.
Under this hierarchy, \(\mathrm{Static}\) indicates that the world state remains locally unchanged.
Among non-static cases, \(\mathrm{Cycle}\) captures periodic recurrence, whereas \(\mathrm{Drift}\) captures changes in the world state that do not advance its active-gap projection.
Second, \name identifies the blocked gap type by comparing \(P_E\) and \(P_A\) among nonempty gap sets.
The type with higher persistence is selected: \(X=E\) indicates blocked epistemic progress, whereas \(X=A\) indicates blocked achievement progress.
Together, the agent dynamics pattern specifies how to escape the current trapping state, while the blocked gap type specifies what progress recovery should restore.

\paragraph{Recovery constraints.}
Based on the diagnosed dynamics pattern and blocked gap type, \name combines a pattern-specific escape constraint with a gap-specific progress constraint:
\begin{equation}
\mathcal{C}_t
=
\mathcal{C}_{\mathrm{pattern}}
\left(\rho_t,\mathcal{W}_t\right)
\cup
\mathcal{C}_{\mathrm{gap}}
\left(X,B_t,\Delta_t\right).
\label{eq:recovery-constraints}
\end{equation}
For \(\mathrm{Static}\), the pattern constraint suppresses the ineffective state--action transition.
For \(\mathrm{Cycle}\), it breaks a recurrent transition identified by \(k_t^*\).
For \(\mathrm{Drift}\), it re-anchors action selection to the active gap.
The gap-specific constraint instead requires new discriminative evidence when \(X=E\) and a task-relevant state change when \(X=A\).
During recovery, the active gap \(\Delta_t\) remains unchanged, while \(\mathcal{C}_t\) is applied as an additional constraint in Eq.~\ref{eq:belief-guided-action}.
After the resulting transition is validated, \name recomputes \(\mathcal{H}_t\).
The recovery constraints are released when \(\mathcal{H}_t>\theta_H\); otherwise, \name updates the diagnosis and corresponding constraints.

%% file: sections/experiments.tex
\section{Experiments}
\label{sec:experiments}

% TODO: Describe settings, baselines, results, ablations, and efficiency analysis.
% Add to main.tex:
% \usepackage{booktabs,multirow}

\paragraph{Benchmarks.}
We evaluate \name on four long-horizon benchmarks covering two complementary forms of agentic decision making.
ALFWorld~\citep{shridhar2020alfworld} and LOCA-Bench~\citep{zeng2026loca} are \emph{goal-directed task-execution} benchmarks, in which the agent repeatedly acts in the environment to bring the current world into a goal-satisfying state.
RCA-100~\citep{cai2026multi} and ClinDiag~\citep{chen2026grounding} are \emph{evidence-seeking diagnosis} benchmarks, where the agent interacts with the environment to acquire discriminative evidence, update competing hypotheses, and ultimately identify the most plausible diagnosis.

\paragraph{Baselines.}
We adopt ReAct~\citep{yao2023react} as the common harness for context-management baselines, using its simple reasoning--action loop to keep the interaction protocol consistent while varying the context supplied for action selection.
\emph{Raw Trajectory} directly provides the complete, chronologically ordered action--observation history.
ACON~\citep{kang2026acon} and PACE~\citep{wei2026pace} manage the growing context through compression and adaptive retention, respectively.
HiAgent~\citep{hu2025hiagent} organizes completed interactions around subgoals and maintains a hierarchical working memory.
LongHorizon-Harness~\citep{ma2026longhorizonharness} maintains an audited task state and coordinates long-horizon execution through a Manage--Execute--Audit architecture.
Together, these baselines cover raw history, compressed and adaptively retained context, hierarchical working memory, and explicitly maintained task state.
LongHorizon-Harness retains its native control architecture, while the remaining baselines operate through the common ReAct harness.

\paragraph{Implementation details.}
We evaluate all methods with three backbone models: \texttt{Qwen3.7-Plus, Kimi-K3, and GLM-5.3.}
Within each backbone setting, all LLM components of \name and all LLM-based components of the baselines use the same evaluated backbone.
All LLM calls use deterministic decoding with temperature 0.
Complete benchmark protocols and budgets, baseline implementations, and method-specific hyperparameters are provided in the appendix~\ref{app:implementation}.

\subsection{Main Results}
\label{sec:main-results}

\begin{table*}[t]
\centering
\caption{Main results across four benchmarks.
We report overall task success for ALFWorld (ALF) and LOCA-Bench (LOCA),
joint accuracy for RCA-100 (RCA), and overall diagnosis accuracy for
ClinDiag (Clin), all in \%.
Best performance is bold. Fine-grained results in Table~\ref{tab:fine-grained-main-results}}
\label{tab:main-results}

\vspace{1mm}

\footnotesize
\setlength{\tabcolsep}{2.2pt}
\renewcommand{\arraystretch}{1.12}

\resizebox{\textwidth}{!}{%
\begin{tabular}{@{}l*{12}{c}@{}}
\toprule

\multirow{2}{*}{\textbf{Method}}
& \multicolumn{4}{c}{%
    \cellcolor{modelgray}%
    \modelicon{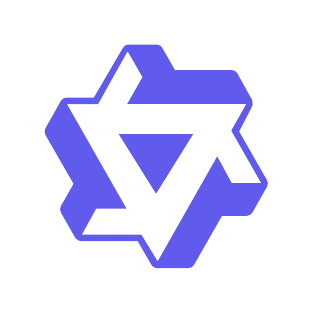}\textbf{Qwen3.7-Plus}}
& \multicolumn{4}{c}{%
    \cellcolor{modelgray}%
    \modelicon{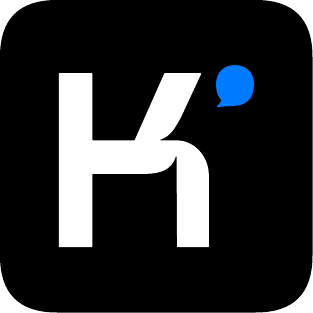}\textbf{Kimi-K3}}
& \multicolumn{4}{c}{%
    \cellcolor{modelgray}%
    \modelicon{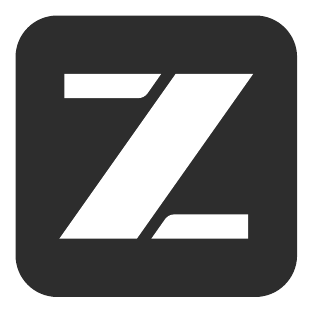}\textbf{GLM-5.3}} \\

\cmidrule(lr){2-5}
\cmidrule(lr){6-9}
\cmidrule(lr){10-13}

& \textbf{ALF}
& \textbf{LOCA}
& \textbf{RCA}
& \textbf{Clin}
& \textbf{ALF}
& \textbf{LOCA}
& \textbf{RCA}
& \textbf{Clin}
& \textbf{ALF}
& \textbf{LOCA}
& \textbf{RCA}
& \textbf{Clin} \\

\specialrule{0.7pt}{0pt}{0.25ex}

Raw Trajectory
& 62.69 & 43.62 & 24.27 & 38.91
& 81.34 & 69.71 & 36.89 & 49.17
& 93.28 & 62.86 & 32.04 & 46.85 \\

ACON
& 66.42 & 49.90 & 27.18 & 39.74
& 81.34 & 70.86 & 40.78 & 38.58
& 94.78 & 65.90 & 33.98 & 46.19 \\

PACE
& 67.91 & 17.33 & 28.16 & 41.89
& 84.33 & 41.52 & 35.92 & 50.17
& 94.78 & 38.29 & 37.86 & 43.38 \\

HiAgent
& 66.42 & 24.57 & 28.16 & 40.07
& 88.81 & 49.52 & 41.75 & 48.51
& 94.78 & 45.90 & 40.78 & 44.87 \\

LongHorizon-Harness
& 72.39 & 52.57 & 26.21 & 40.56
& 91.79 & 67.62 & 38.83 & 51.82
& 95.52 & 64.38 & 31.07 & 45.53 \\

\specialrule{0.35pt}{0.35ex}{0.20ex}

\ColorFirstCell{posmain}{\PoSMethod}
& \ColorMiddleCell{posmain}{\textbf{88.81}}
& \ColorMiddleCell{posmain}{\textbf{56.38}}
& \ColorMiddleCell{posmain}{\textbf{38.83}}
& \ColorMiddleCell{posmain}{\textbf{45.03}}
& \ColorMiddleCell{posmain}{\textbf{94.03}}
& \ColorMiddleCell{posmain}{\textbf{74.29}}
& \ColorMiddleCell{posmain}{\textbf{51.46}}
& \ColorMiddleCell{posmain}{\textbf{54.80}}
& \ColorMiddleCell{posmain}{\textbf{97.01}}
& \ColorMiddleCell{posmain}{\textbf{70.86}}
& \ColorMiddleCell{posmain}{\textbf{49.51}}
& \ColorLastCell{posmain}{\textbf{52.15}} \\

\ColorFirstCell{posablation}{%
    \quad - Consistency Validation%
}
& \ColorMiddleCell{posablation}{73.88}
& \ColorMiddleCell{posablation}{44.57}
& \ColorMiddleCell{posablation}{31.07}
& \ColorMiddleCell{posablation}{44.54}
& \ColorMiddleCell{posablation}{91.79}
& \ColorMiddleCell{posablation}{67.43}
& \ColorMiddleCell{posablation}{48.54}
& \ColorMiddleCell{posablation}{54.47}
& \ColorMiddleCell{posablation}{94.78}
& \ColorMiddleCell{posablation}{62.67}
& \ColorMiddleCell{posablation}{47.57}
& \ColorLastCell{posablation}{51.49} \\

\ColorFirstCell{posablation}{%
    \quad - Trapping Diagnosis%
}
& \ColorMiddleCell{posablation}{73.13}
& \ColorMiddleCell{posablation}{48.19}
& \ColorMiddleCell{posablation}{33.98}
& \ColorMiddleCell{posablation}{42.38}
& \ColorMiddleCell{posablation}{88.81}
& \ColorMiddleCell{posablation}{69.52}
& \ColorMiddleCell{posablation}{46.60}
& \ColorMiddleCell{posablation}{50.83}
& \ColorMiddleCell{posablation}{92.54}
& \ColorMiddleCell{posablation}{65.71}
& \ColorMiddleCell{posablation}{42.72}
& \ColorLastCell{posablation}{48.51} \\

\bottomrule
\end{tabular}%
}
\end{table*}

\paragraph{(1) Existing context management does not consistently improve upon raw trajectories.}
Table~\ref{tab:main-results} reveals substantial variation in baseline performance across tasks and backbones.
PACE remains competitive on ALFWorld and RCA-100, but falls \(24.57\)--\(28.19\) points below Raw Trajectory on LOCA-Bench.
LongHorizon-Harness performs well on LOCA-Bench, where explicit requirements and generated artifacts support intermediate verification, but delivers less consistent gains on RCA-100 and ClinDiag, as progress depends on refining uncertain diagnostic hypotheses rather than producing directly verifiable artifacts.
On ClinDiag with \texttt{GLM-5.3}, none of the context-management baselines outperforms Raw Trajectory.
These reversals suggest that additional compression and organization do not consistently benefit modern LLMs with strong long-context capabilities, as information loss may offset the benefits of reducing redundancy.

\paragraph{(2) \name consistently outperforms existing methods across diverse tasks and backbones.}
In contrast, \name achieves the highest overall performance on all four benchmarks under all three backbones.
Compared with the strongest baseline with the same backbone, \name achieves relative gains of up to \(22.68\%\) on ALFWorld, \(7.53\%\) on LOCA-Bench, \(37.89\%\) on RCA-100 joint accuracy, and \(11.31\%\) on ClinDiag.
The gains span both execution and diagnosis tasks, indicating that they are not tied to a particular task structure.
These results highlight the importance of maintaining an accurate, task-conditioned belief about the current world state and unresolved requirements, beyond retaining and organizing interaction history.

\paragraph{(3) Effective belief modeling requires maintaining both consistency and health.}
Removing Consistency Validation lowers performance across all four benchmarks to varying degrees, with losses reaching \(14.93\) points on ALFWorld and \(11.81\) points on LOCA-Bench, compared with \(0.33\)--\(0.66\) points on ClinDiag.
This pattern is consistent with the greater risk of inconsistent belief updates when complex interactions require the agent to reconcile changing states and observations across steps.
Removing Trapping Diagnosis and its associated recovery also degrades performance across all four benchmarks, including losses of \(4.85\)--\(6.79\) points on RCA-100 and \(2.65\)--\(3.97\) points on ClinDiag.
Recovery complements validation by redirecting actions toward unresolved requirements when continued interaction no longer advances the task.
Together, these results demonstrate the value of maintaining both belief consistency and task progress, beyond constructing an explicit representation.

\subsection{Belief Trapping Analysis}
\label{sec:belief-trapping-analysis}

\begin{figure}[ht]
    \centering
    \includegraphics[width=\linewidth]{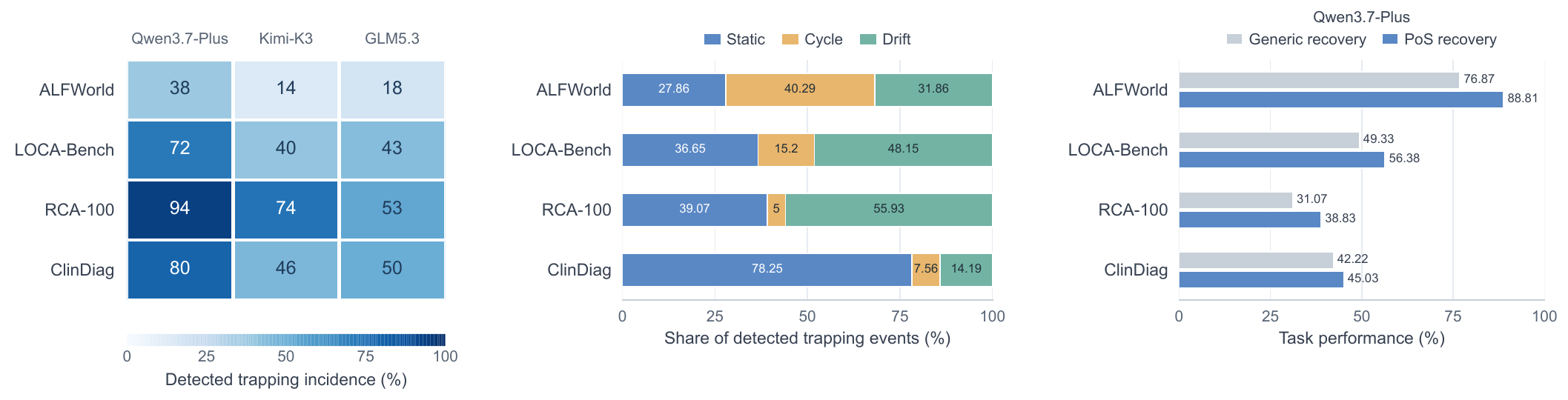}
    \caption{\textbf{Belief Trapping and recovery across benchmarks.}
\textbf{Left:} Percentage of episodes with detected trapping under each backbone.
\textbf{Middle:} Distribution of trapping patterns by benchmark.
\textbf{Right:} Final task performance over all evaluated cases with Qwen3.7-Plus, comparing \name's factorized recovery with a generic recovery prompt.}
    \label{fig:belief-trapping-analysis}
\end{figure}

Figure~\ref{fig:belief-trapping-analysis} summarizes detected trapping incidence (left), the distribution of trapping patterns (middle), and the effect of recovery strategies after trapping is detected (right).
It reveals that trapping is common across different settings, although stronger backbones generally exhibit lower incidence.
However, lower trapping incidence does not necessarily imply higher final accuracy: on RCA-100, \texttt{Kimi-K3} exhibits more trapping than \texttt{GLM-5.3} (\(74\%\) vs.\ \(53\%\)) but achieves higher Joint accuracy (Table~\ref{tab:main-results}).
The dominant trapping pattern also varies across benchmarks, with cycles most frequent in ALFWorld, drift in LOCA-Bench and RCA-100 (\(55.93\%\)), and static stagnation in ClinDiag (\(78.25\%\)).
These differences reflect how unproductive interactions affect belief transition.
Embodied actions may revisit prior states, multisource evidence may update beliefs without advancing the active gap, and non-discriminative clinical investigations may leave diagnostic beliefs unchanged.
To assess the value of factorized recovery, we compare \name with a variant that replaces factorized recovery constraints with a generic prompt asking the agent to recover.
Figure~\ref{fig:belief-trapping-analysis} (right) shows that \name's recovery strategy outperforms this variant across all four benchmarks.
These results demonstrate that effective recovery goes beyond interrupting unproductive behavior, with factorized recovery tailored to the trapping dynamics and blocked gap type.

\subsection{Cost--Performance Tradeoff}
\label{sec:cost-performance}

\begin{table*}[ht]
\centering
\caption{Cost--performance tradeoff on RCA-100.
We report joint accuracy (\%) and mean token consumption per episode in thousands (K), averaged over all 103 cases.
Token counts include input and output tokens for each component.}
\label{tab:cost-performance}

\footnotesize
\setlength{\tabcolsep}{5pt}
\renewcommand{\arraystretch}{1.12}

\begin{tabular}{@{}llccccc@{}}
\toprule
\multirow{2}{*}{\textbf{Backbone}} &
\multirow{2}{*}{\textbf{Method}} &
\multirow{2}{*}{\textbf{Joint (\%)}} &
\multicolumn{4}{c}{\textbf{Tokens per episode (K)}} \\
\cmidrule(lr){4-7}
& & & Task Agent & Belief & Sentinel & \textbf{Total} \\
\midrule

\multirow{4}{*}{Qwen3.7-Plus}
& Raw Trajectory
& 24.27 & 355.70 & 0.00 & 0.00 & 355.70 \\
& \name{} (Ours)
& \textbf{38.83} & 281.20 & 696.69 & 822.24 & 1800.13 \\
& \quad - Consistency Validation
& 31.07 & 299.82 & 733.41 & 125.30 & 1158.53 \\
& \quad - Trapping Diagnosis
& 33.98 & 327.40 & 657.47 & 649.11 & 1633.98 \\

\bottomrule
\end{tabular}
\end{table*}

% Table~\ref{tab:cost-performance} compares diagnostic performance and token consumption on RCA-100 across three backbones.
% Compared with Raw Trajectory, \name achieves relative improvements of \(39.5\%\)--\(60.0\%\) in joint accuracy, while reducing token consumption by the Task Agent by \(20.9\%\)--\(36.6\%\).
% These gains come with higher overall consumption, amounting to \(5.06\)--\(7.04\times\) the tokens used by Raw Trajectory.
% The additional computation is devoted to constructing and validating the belief that guides subsequent decisions, rather than increasing the tokens processed by the Task Agent itself.
% Thus, \name trades the computational cost of explicit belief maintenance for more accurate diagnosis, with a smaller token footprint for the Task Agent.

Table~\ref{tab:cost-performance} examines the cost--performance tradeoff on RCA-100 with \texttt{Qwen3.7-Plus}.
Compared with Raw Trajectory, \name improves joint accuracy from \(24.27\%\) to \(38.83\%\), while reducing Task Agent token consumption by \(20.9\%\).
However, total consumption increases to \(5.06\times\), as additional computation is devoted to belief construction and maintenance.
Removing Consistency Validation substantially reduces Sentinel overhead and saves \(35.6\%\) of total tokens, but sacrifices \(7.76\) points in joint accuracy.
By contrast, removing Trapping Diagnosis and its associated recovery saves only \(9.2\%\) of total tokens, while increasing Task Agent consumption and reducing joint accuracy by \(4.85\) points.
Both ablations therefore reduce overall cost but leave the Task Agent consuming more tokens with less accurate final diagnoses.
These results highlight the value of investing additional computation in belief maintenance to improve subsequent decisions, rather than minimizing token consumption alone.

\subsection{Scaling under Context Growth}
\label{sec:context-scaling}

\begin{figure}[ht]
    \centering
    \includegraphics[width=\linewidth]{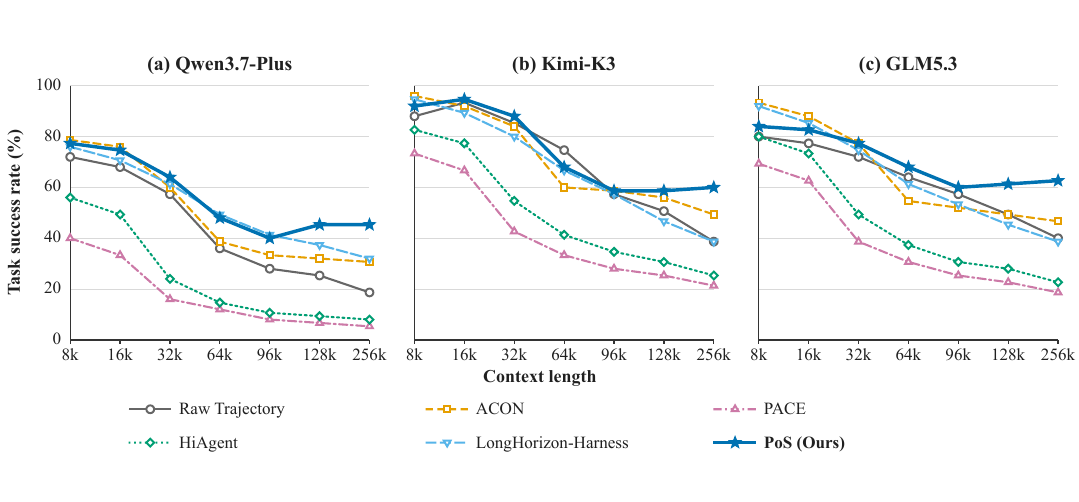}
    \caption{\textbf{Success rates on LOCA-Bench as context grows from 8K to 256K across backbones.}}
    \label{fig:context-scaling}
\end{figure}

% Figure~\ref{fig:context-scaling} compares task success on LOCA-Bench as environment descriptions grow from 8K to 256K across three backbones, and performance generally declines with context length.
% ACON compresses accumulated interaction history into a summary whenever the context exceeds a predefined threshold.
% At 256K, it outperforms Raw Trajectory under all three backbones, suggesting that reducing redundancy can offset some of the information loss introduced by compression as interaction histories expand.
% Meanwhile, \name{} maintains broadly stable performance from 96K to 256K, exceeding the strongest baseline by \(10.67\)--\(16.00\) points at 256K.
% A plausible explanation is that the task state needed for action selection need not grow in proportion to the surrounding environmental information.
% By integrating relevant observations into an existing world state and unresolved requirements, \name{} reduces the need to repeatedly reconstruct the current task state from a growing history.
% Continual validation and progress monitoring help prevent unsupported updates and unproductive interactions from undermining this state as execution proceeds.
% The agent can therefore base subsequent decisions on an updated task state rather than repeatedly reconcile dispersed or outdated observations, providing a mechanism for greater resilience to context growth.
% Together, these results suggest that effective context management should balance reducing historical redundancy with preserving and updating the information needed for an actionable task state.

Figure~\ref{fig:context-scaling} compares task success on LOCA-Bench as environment descriptions grow from 8K to 256K across three backbones.
Although performance generally declines with context length, ACON shows a more gradual decline than Raw Trajectory at longer contexts.
By compressing accumulated interaction history whenever the context exceeds a predefined threshold, ACON reduces redundancy, whose increasing burden can make compression beneficial despite potential information loss.
Meanwhile, \name{} maintains broadly stable performance from 96K to 256K, exceeding the strongest baseline by \(10.67\)--\(16.00\) points at 256K.
By integrating new observations into an existing world state and unresolved requirements, \name{} grounds each decision in the current task conditions rather than requiring the agent to reconstruct them from an increasingly long interaction history.
This incremental maintenance reduces repeated interpretation of dispersed or outdated observations, while consistency validation and progress monitoring help prevent unsupported updates and stalled interactions from persisting.
Together, these results suggest that effective context management should balance reducing historical redundancy with maintaining a reliable and actionable understanding of the current world.

%% file: sections/conclusion.tex
\section{Conclusion}
\label{sec:conclusion}

In this work, we showed that explicit belief construction alone is insufficient for reliable long-horizon decision-making.
We proposed \name{}, a framework that constructs task-conditioned belief states and continually maintains them through consistency validation and factorized recovery from \emph{Belief Trapping}.
Experiments across four benchmarks and three LLM backbones demonstrate consistent improvements in execution and diagnosis, together with resilience to context growth.
Although \name{} uses more tokens, the resulting gains demonstrate the potential of investing inference-time computation in belief maintenance.
Reducing this overhead while preserving the gains in agent performance is an important direction for future work.

%% file: sections/appendix.tex
\appendix

\clearpage
\setcounter{tocdepth}{2}
\renewcommand{\contentsname}{Contents}
\tableofcontents
\clearpage

% =========================================================
% Appendix A: Notation
% =========================================================
\section{Notation Summary}
\label{app:notation}

Table~\ref{tab:notation-summary} summarizes the notation used throughout the methodology and supplementary material.

\begin{table*}[ht]
\centering
\caption{Summary of the principal notation used in \name.}
\label{tab:notation-summary}
\footnotesize
\setlength{\tabcolsep}{5pt}
\renewcommand{\arraystretch}{1.12}
\begin{tabularx}{\textwidth}{@{}p{0.17\textwidth}p{0.34\textwidth}X@{}}
\toprule
\textbf{Symbol} & \textbf{Meaning} & \textbf{Domain / Notes} \\
\midrule
\multicolumn{3}{@{}l}{\textbf{Environment and interaction}} \\
$s_t$
& Latent world state at step $t$
& Not directly observable by the agent. \\
$h_t$
& Interaction history at step $t$
& Contains previous actions and observations. \\
$a_t,\ o_{t+1}$
& Agent action and resulting observation
& The environment returns $o_{t+1}$ after action $a_t$. \\
$T$
& Maximum interaction budget
& Defined separately for each benchmark. \\
\midrule
\multicolumn{3}{@{}l}{\textbf{Belief representation and update}} \\
$b_t(s)$
& Posterior belief over the latent world state
& $b_t(s)=P(s_t=s\mid h_t)$ in the POMDP formulation. \\
$B_t$
& Task-conditioned belief state
& $B_t=(W_t,G,\Delta_t^E,\Delta_t^A)$. \\
$W_t$
& Task-relevant world state
& Structured through Entities, States, and Relations. \\
$G$
& Task goal
& Specifies the objective and completion criteria. \\
$\Delta_t^E,\ \Delta_t^A$
& Epistemic and achievement gap sets
& Information to clarify and goal conditions to achieve. \\
$\Delta_t$
& Active gap at step $t$
& $\Delta_t\in\Delta_t^E\cup\Delta_t^A$; retained until resolved. \\
$\widetilde{B}_{t+1}$
& Candidate belief update
& Validated before being committed as $B_{t+1}$. \\
$\mathcal{U}$
& Candidate belief-update operator
& $\widetilde{B}_{t+1}=\mathcal{U}(B_t,a_t,o_{t+1})$. \\
$\pi$
& Action-selection policy
& Conditions on $B_t$, $\Delta_t$, and $\mathcal{C}_t$. \\
\midrule
\multicolumn{3}{@{}l}{\textbf{Progress and belief health}} \\
$c_t(x)$
& Confidence assigned to inferred content $x$
& Independent confidence scores for goal-relevant inferred States and Relations, without normalization across records. \\
$d_t^{\mathrm{diag}}$
& Diagnostic progress measure
& Total variation between $c_t$ and $c_{t-1}$. \\
$u_t$
& Progress label for $B_{t-1}\rightarrow B_t$
& $u_t\in\{0,1\}$; $1$ indicates recorded task progress. \\
$\mathcal{W}_t$
& Window of recent validated transitions
& $\mathcal{W}_t=\{t-K+1,\ldots,t\}$. \\
$P_X,\ S_t,\ R_t$
& Gap persistence, progress stagnation, and belief recurrence
& Window-level signals used to estimate belief health. \\
$W_\tau^\Delta$
& World state projected onto the active gap
& Projects $W_\tau$ onto the current active gap $\Delta_t$. \\
$d_J,\ d_W$
& Jaccard distance and world state distance
& $d_W$ averages $d_J$ over Entity, State, and Relation sets. \\
$\mathcal{H}_t$
& Belief health score
& $\mathcal{H}_t=1-\max_{X\in\{E,A\}}[P_X\max(S_t,R_t)]$. \\
$K,\ L$
& Window size and maximum recurrence lag
& Recurrence is evaluated over lags $1\leq\ell\leq L<K$. \\
$\epsilon,\ \epsilon_W$
& Diagnostic progress and state similarity thresholds
& Used to identify confidence changes and recurring world states. \\
$\theta_H,\ \theta_R$
& Health and cycle-detection thresholds
& Trapping requires $\mathcal{H}_t\leq\theta_H$; cycle diagnosis uses $R_t\geq\theta_R$. \\
\midrule
\multicolumn{3}{@{}l}{\textbf{Factorized diagnosis and recovery}} \\
$\rho_t$
& Diagnosed trapping dynamics
& $\rho_t\in\{\text{Static},\text{Cycle},\text{Drift}\}$. \\
$k_t^*$
& Dominant recurrence lag
& The lag underlying $R_t$; cycle diagnosis additionally requires $k_t^*>1$. \\
$X,\ X_t$
& Gap type and its selection at step $t$
& $X\in\{E,A\}$; $X_t$ denotes the selected type(s) in Algorithm~\ref{alg:pos-overall}. \\
$\mathcal{C}_t$
& Recovery constraints for action selection
& Combines $C_{\mathrm{pattern}}$ and $C_{\mathrm{gap}}$; empty during normal interaction. \\
\bottomrule
\end{tabularx}
\end{table*}

% =========================================================
% Appendix B: Supplementary method details
% =========================================================
\section{Supplementary Method Details}
\label{app:method-details}

\subsection{End-to-End Workflow of \name}
\label{app:end-to-end-workflow}
% 内容：给出从初始化到终止的完整可执行流程，以及正文省略的 prompts 和 schema。
% 作用：正文解释方法设计，附录负责让读者能够实现该方法。

% 内容：
% 1. 给出 \name 的完整算法伪代码。
% 2. 从任务目标和初始 observation 构造 \(B_0\)。
% 3. 依次表示 active gap selection、action selection、environment interaction、
%    belief update、Sentinel validation、health estimation、trapping diagnosis
%    和 recovery。
% 4. 明确普通交互、触发 recovery 和任务终止三种控制分支。
%
% 为什么要放：
% 正文中的三个模块分别介绍，但读者仍需要一个端到端算法理解它们在每一步
% 的调用顺序及数据依赖。
%
% 建议形式：
% 使用 algorithmic 或 algorithm2e 给出一个主算法。
% 尽量复用正文符号，避免在伪代码中重新引入新的状态变量。

Algorithm~\ref{alg:pos-overall} summarizes the end-to-end inference workflow of \name.
It follows the notation introduced in Section~\ref{sec:methodology} and abstracts away implementation-specific details while preserving the main control flow and data dependencies of the framework.
In particular, the Active Gap is maintained across consecutive steps until it is resolved, and trapping is assessed only after the current belief transition has been validated and incorporated into Belief Health.
Any recovery constraint triggered by the resulting health state therefore guides action selection from the subsequent interaction step.

\begin{algorithm}[ht]
\caption{Overall Inference Workflow of \name}
\label{alg:pos-overall}
\small
\begin{algorithmic}[1]

\Require Task goal $G$, initial observation $o_0$, interaction budget $T$.
\Require Window size $K$, maximum lag $1 \leq L < K$, thresholds $\epsilon$, $\epsilon_W$, $\theta_H$, and $\theta_R$.
\Ensure Environment result or termination upon budget exhaustion.

\State $(B_0,\Delta_0) \gets \Call{InitializeBelief}{G,o_0}$.
\Comment{$B_0=(W_0,G,\Delta_0^E,\Delta_0^A)$}
\State $\mathcal{C}_0 \gets \emptyset$.

\For{$t=0,1,\ldots,T-1$}

    \State $a_t \sim \pi(\cdot \mid B_t,\Delta_t,\mathcal{C}_t)$.
    \Comment{$\mathcal{C}_t=\emptyset$ during normal interaction}

    \State $o_{t+1} \gets \Call{EnvironmentStep}{a_t}$.

    \State $\widetilde{B}_{t+1}
    \gets \mathcal{U}(B_t,a_t,o_{t+1})$.
    \Comment{Candidate belief update}

    \State $B_{t+1}
    \gets \Call{BeliefSentinel}
    {B_t,\widetilde{B}_{t+1},a_t,o_{t+1}}$.
    \Comment{Audit and revise before commitment}

    \If{the task is diagnostic}
        \State $d_{t+1}^{\mathrm{diag}}
        \gets \frac{1}{2}\sum_x
        \left|c_{t+1}(x)-c_t(x)\right|$.
        \State $u_{t+1}
        \gets \mathbb{I}\!\left[d_{t+1}^{\mathrm{diag}}>\epsilon\right]$.
    \Else
        \State $u_{t+1}
        \gets \Call{LabelExecutionProgress}{B_t,B_{t+1},\Delta_t}$.
        \Comment{Belief Sentinel judgment}
    \EndIf

    \If{\Call{IsTerminal}{$o_{t+1}$}}
        \State \Return \Call{EnvironmentResult}{}.
    \EndIf

    \State $\Delta_{t+1}
    \gets \Call{AdvanceActiveGap}{B_{t+1},\Delta_t}$.
    \Comment{Keep $\Delta_t$ unless it is resolved}

    \State $\mathcal{C}_{t+1} \gets \emptyset$.

    \If{$t+1 \geq K$}
        \State $\mathcal{W}_{t+1}
        \gets \{t-K+2,\ldots,t+1\}$.

        \State Compute $P_E$, $P_A$, $S_{t+1}$, and $R_{t+1}$
        over $\mathcal{W}_{t+1}$ using Eq.~\ref{eq:belief-health}.
        \Statex \hspace{\algorithmicindent}
        Project each $W_\tau$ onto the current active gap $\Delta_{t+1}$.

        \State $\mathcal{H}_{t+1}
        \gets 1-\max_{X\in\{E,A\}}
        \left[P_X\max(S_{t+1},R_{t+1})\right]$.

        \If{$\mathcal{H}_{t+1}\leq\theta_H$}

            \State $\rho_{t+1}
            \gets \Call{ClassifyDynamics}
            {\{B_\tau\}_{\tau\in\mathcal{W}_{t+1}},\Delta_{t+1}}$.
            \Statex \hspace{\algorithmicindent}
            Apply the \textsc{Static}, \textsc{Cycle}, and \textsc{Drift}
            conditions in order.

            \State $X_{t+1}
            \gets \Call{SelectBlockedGapType}{P_E,P_A,B_{t+1}}$.
            \Statex \hspace{\algorithmicindent}
            Select the higher-persistence type among nonempty gap sets.
            On a tie, prioritize epistemic gaps for diagnostic tasks
            and achievement gaps for execution tasks.
            
            \State $\mathcal{C}_{t+1}
            \gets
            \mathcal{C}_{\mathrm{pattern}}
            (\rho_{t+1},\mathcal{W}_{t+1})
            \cup
            \mathcal{C}_{\mathrm{gap}}
            (X_{t+1},B_{t+1},\Delta_{t+1})$.

        \EndIf
    \EndIf

\EndFor

\State \Return \textsc{BudgetExhausted}.

\end{algorithmic}
\end{algorithm}

\subsection{Belief Trapping Detection and Recovery}
\label{app:trapping-details}

\paragraph{Step-level progress assessment.}
For diagnostic tasks, \(c_t(x)\in[0,1]\) denotes the confidence assigned to a goal-relevant inferred State or Relation.
These values are maintained independently without normalization across records.
The progress measure in Eq.~\ref{eq:diagnostic-progress} aggregates absolute confidence changes between consecutive validated beliefs, so both increases and decreases contribute to the progress label when the aggregate change exceeds \(\epsilon\).
For execution tasks, the Belief Sentinel evaluates progress relative to the active gap using the criteria in Section~\ref{sec:trapping-aware-recovery}.

\paragraph{Window-based health estimation.}
Trapping detection begins once \(K\) validated transitions are available.
At each step \(t\), historical world states within \(\mathcal{W}_t\) are projected onto the same current active gap \(\Delta_t\), selecting relevant records through their existing relations.
The comparison underlying \(d_W\) includes confidence changes even when descriptive content remains unchanged.
When the active gap changes, the existing window is retained and its historical world states are projected onto the new gap.
The resulting persistence, stagnation, and recurrence signals are combined according to Eq.~\ref{eq:belief-health}, with trapping detected when \(\mathcal{H}_t\leq\theta_H\).

\paragraph{Factorized trapping diagnosis.}
Once trapping is detected, the \textsc{Static}, \textsc{Cycle}, and \textsc{Drift} conditions are evaluated in order.
If none is satisfied, \name uses a generic recovery prompt rather than assigning a trapping pattern.
The blocked gap type is selected by comparing persistence across nonempty gap sets.
When persistence is equal, diagnostic tasks prioritize epistemic gaps, whereas execution tasks prioritize achievement gaps.

\paragraph{Recovery constraint composition.}
The diagnosed pattern and selected gap type are incorporated into a recovery prompt together with the relevant Entities and the active gap.
The resulting constraints address both the unproductive behavior and the unresolved requirement, and guide subsequent action selection while the active gap remains unchanged.
After each validated transition, \name recomputes belief health and releases the constraints when \(\mathcal{H}_t>\theta_H\).
Otherwise, it updates the diagnosis and corresponding recovery constraints.

\subsection{Design Rationale for Belief Trapping}
\label{app:trapping-rationale}

\paragraph{Choice of trapping signals.}
The three signals capture complementary aspects of unproductive interaction.
Gap Persistence ties the assessment to requirements that remain unresolved, while Progress Stagnation measures the absence of recorded task progress.
Belief Recurrence complements these signals by identifying repeated belief configurations that may occur despite positive step-level progress labels.
Behavioral statistics such as action repetition or interaction count do not directly establish whether the task is advancing.
Repeated tool calls may acquire relevant information, whereas different actions may leave the same requirement unresolved.
Likewise, confidence levels alone do not indicate progress, as high confidence can persist during stalled investigation and low confidence can accompany useful evidence gathering.
The selected signals instead connect unresolved requirements with changes and recurring patterns in the maintained belief.

\paragraph{Aggregation of trapping signals.}
We design the health score to capture stagnation or recurrence associated with persistent unresolved requirements, without treating gap persistence alone as sufficient evidence of trapping.
Let \(p_t=\max_{X\in\{E,A\}}P_X\) denote the largest gap persistence within the current window.
A geometric mean can miss trapping because stagnation and recurrence need not occur together.
Specifically, \(\mathcal{H}_t^{\mathrm{geo}}=1-(p_tS_tR_t)^{1/3}\) yields \(\mathcal{H}_t^{\mathrm{geo}}=1\) whenever either \(S_t\) or \(R_t\) is zero, regardless of how high the other signals are.
An agent may make no progress without revisiting earlier beliefs, or repeatedly return to earlier beliefs while confidence changes produce positive progress labels.
In either case, the absence of one signal prevents the geometric mean from identifying trapping.
By contrast, \(\mathcal{H}_t^{\mathrm{max}}=1-\max(p_t,S_t,R_t)\) can incorrectly identify trapping solely because unresolved requirements persist.
When \(p_t=1\), it yields \(\mathcal{H}_t^{\mathrm{max}}=0\) even if \(S_t=R_t=0\), meaning that every transition records progress and no belief recurrence is detected.
A gap may remain unresolved throughout the window while the agent makes intermediate progress toward resolving it.
The maximum formulation therefore confuses an unfinished requirement with stalled progress.
\name therefore uses
\begin{equation}
    \mathcal{H}_t
    =
    1-p_t\max(S_t,R_t),
    \label{eq:health-aggregation-rationale}
\end{equation}
which weights the stronger stagnation or recurrence signal by the persistence of unresolved requirements.
This preserves sensitivity to either form of unproductive belief evolution without treating gap persistence alone as sufficient evidence of trapping.
Table~\ref{tab:health-aggregation-cases} illustrates these distinctions through three boundary configurations.

\begin{table*}[ht]
\centering
\caption{Health scores under three illustrative boundary configurations.
Lower values indicate stronger trapping signals.}
\label{tab:health-aggregation-cases}
\small
\setlength{\tabcolsep}{6pt}
\renewcommand{\arraystretch}{1.15}
\begin{tabularx}{\textwidth}{@{}Xcccc@{}}
\toprule
\textbf{Boundary case}
& \(\boldsymbol{(p_t,S_t,R_t)}\)
& \(\mathcal{H}_t^{\mathrm{geo}}\)
& \(\mathcal{H}_t^{\mathrm{max}}\)
& \(\mathcal{H}_t\) \\
\midrule
Stagnation without recurrence
& \((1,1,0)\) & \(1\) & \(0\) & \(0\) \\
Recurrence despite positive progress labels
& \((1,0,1)\) & \(1\) & \(0\) & \(0\) \\
Progress without gap closure
& \((1,0,0)\) & \(1\) & \(0\) & \(1\) \\
\bottomrule
\end{tabularx}
\end{table*}

% =========================================================
% Appendix C: Benchmark details
% =========================================================
\section{Benchmark Details and Evaluation Protocols}
\label{app:benchmarks}
% 内容：补充正文因篇幅限制而省略的 benchmark 定义、数据组成、
% 交互方式、评价指标和终止条件。
% 作用：使读者能够准确理解主表中每一列的含义，并复现相同的评测协议。

\subsection{Task Taxonomy}
% 内容：
% 1. 给出 ALFWorld、LOCA-Bench、RCA-100 和 ClinDiag 的任务分类。
% 2. 列出每个聚合类别所包含的原始任务类型。
% 3. 解释分类依据，例如任务目标、主要环境操作或证据获取方式。
%
% 为什么要放：
% 正文只报告聚合类别结果。如果没有完整映射，读者无法判断各类别包含
% 哪些任务，也无法理解类别间的性能差异。
%
% 建议形式：
% 使用 itemize 或一张“Benchmark--Category--Original Task Type”映射表。
% 当前已有的 ALFWorld 和 LOCA-Bench 分类可以直接放在这里，并补充
% RCA-100 与 ClinDiag 是否存在可用的细粒度类别。
For concise reporting, we organize each benchmark according to its native task structure or evaluation dimensions.
ALFWorld and LOCA-Bench are grouped by task type, while RCA-100 and ClinDiag follow the diagnostic targets and dataset subsets defined by their respective benchmarks.

\subsubsection{ALFWorld}

\begin{itemize}
    \item \textbf{Place}: Pick \& Place; Pick Two \& Place.
    \item \textbf{Transform}: Clean \& Place; Heat \& Place; Cool \& Place.
    \item \textbf{Inspect}: Examine in Light.
\end{itemize}

\subsubsection{LOCA-Bench}

\begin{itemize}
    \item \textbf{Cross-source Workflows}: Canvas Arrange Exam; Course Assistant; Apply PhD Email; Set Conference/CR Deadline; Payable Invoice Checker.
    \item \textbf{Structured Analytics}: A/B Testing; Academic Warning; Excel Market Research; Machine Operating; NHL B2B Analysis.
    \item \textbf{Environment Operations}: Canvas List Test; Filter Low-Selling Products; Update Material Inventory; WooCommerce New Welcome; WooCommerce Stock Alert.
\end{itemize}

\subsubsection{RCA-100}

\begin{itemize}
    \item \textbf{Fault Triage (FT)}: Identify the type of failure responsible for the observed incident.
    \item \textbf{Root-Cause Localization (RCL)}: Locate the entity at which the root cause occurs.
    \item \textbf{Joint}: Require both the fault type and the root-cause entity to be correctly identified.
\end{itemize}

\subsubsection{ClinDiag}

\begin{itemize}
    \item \textbf{Common}: Real-world clinical cases collected from published case reports across multiple medical specialties.
    \item \textbf{Rare}: Clinical cases covering diverse categories of rare diseases.
\end{itemize}

\subsection{Dataset Statistics and Splits}
\label{app:dataset-statistics}
% 内容：
% 1. 报告每个 benchmark 的原始规模和实际评测 case 数量。
% 2. 说明使用的 split，例如 validation、test 或官方公开子集。
% 3. 报告每个任务类别中的 case 数量。
% 4. 说明是否排除了无效、缺失或无法执行的 cases，以及排除标准。
% 5. 对 LOCA-Bench，报告不同 context length 下的 case 数量是否一致。
%
% 为什么要放：
% 聚合准确率会受到类别规模影响。读者需要知道 Overall 是在怎样的
% 样本分布上计算的，也需要排除因数据筛选造成的结果偏差。
%
% 建议形式：
% 使用一张统计表，列出 Benchmark、Split、Category、Number of Cases、
% Number of Runs。若有 case filtering，再单独用一段说明筛选流程。

Table~\ref{tab:dataset-statistics} summarizes the evaluation splits and
numbers of cases used in our experiments.

\begin{table*}[ht]
\centering
\caption{Dataset statistics and evaluation splits.}
\label{tab:dataset-statistics}
\small
\setlength{\tabcolsep}{6pt}
\begin{tabular}{@{}lllr@{}}
\toprule
\textbf{Benchmark} &
\textbf{Evaluation Split} &
\textbf{Task Category} &
\textbf{\# Cases} \\
\midrule

ALFWorld
& \texttt{valid\_unseen}
& Place
& 41 \\

&
& Transform
& 75 \\

&
& Inspect
& 18 \\

&
& \textbf{Total}
& \textbf{134} \\

\midrule

LOCA-Bench
& Official benchmark
& Cross-source Workflows
& 175 \\

&
& Structured Analytics
& 175 \\

&
& Environment Operations
& 175 \\

&
& \textbf{Total}
& \textbf{525} \\

\midrule

RCA-100
& Full benchmark
& All diagnostic cases
& 103 \\

\midrule

ClinDiag
& Sampled evaluation subset
& Common
& 302 \\

&
& Rare
& 302 \\

&
& \textbf{Total}
& \textbf{604} \\

\bottomrule
\end{tabular}
\end{table*}

For ALFWorld, we use the standard \texttt{valid\_unseen} split containing
134 tasks in previously unseen environments.
The split contains 41 Place, 75 Transform, and 18 Inspect tasks, and the same
cases are used for every method and backbone.

LOCA-Bench contains 15 seed tasks instantiated at seven environment-description
lengths: 8K, 16K, 32K, 64K, 96K, 128K, and 256K tokens.
For each task and target length, five random environment seeds generate
distinct environment states while preserving the task definition.
This produces 75 cases at each context length and 525 cases in total.
Each of our three task categories contains five seed tasks and therefore
175 cases.
The environment seeds control benchmark instantiation and should not be
confused with repeated model runs.

For RCA-100, we evaluate all 103 cases in the released benchmark.
Every case is scored on fault triage (FT), root-cause localization (RCL), and
their joint correctness, so these targets share the same underlying set of
diagnostic cases.

The complete ClinDiag benchmark additionally contains an access-controlled
Emergency subset derived from MIMIC-IV-Ext, which was unavailable under our
current data-access setting and is therefore excluded.
Because the available Common pool is substantially larger than the Rare
subset, we randomly sample 302 Common cases to match all 302 available Rare
cases.
This yields a balanced evaluation set of 604 cases, prevents the Common
subset from dominating the aggregate result, and keeps the evaluation
computationally tractable.
The Common subset is sampled once before evaluation and is held fixed across
all methods and backbones.

\subsection{Interactive Evaluation Protocols}
% 内容：
% 1. 说明四个 benchmark 在本文中如何被组织为多轮 Agent--Environment 交互。
% 2. 对 ALFWorld 和 LOCA-Bench，说明环境 observation、可用动作和成功反馈。
% 3. 对 RCA-100，说明 Agent 如何查询 telemetry，以及如何输出 FT 和 RCL。
% 4. 对 ClinDiag，说明初始可见的临床信息、可请求的检查或证据，
%    以及 Agent 何时提交最终诊断。
% 5. 如果本文修改了 benchmark 的原始交互接口，需要明确描述修改内容。
%
% 为什么要放：
% RCA-100 和 ClinDiag 属于 evidence-seeking diagnosis，而 ALFWorld 和
% LOCA-Bench 属于 goal-directed execution。二者的交互语义不同，不能只用
% 一句“均采用 ReAct”代替具体协议。
%
% 建议形式：
% 每个 benchmark 使用一个短段落，并给出一个简化的交互示例。
% 若诊断 benchmark 是由静态数据改造成交互形式，建议增加一张流程图，
% 展示 Initial Evidence--Evidence Request--Observation--Final Diagnosis。

All methods interact with each benchmark through the same benchmark-specific interface.
The context-management baselines operate within the shared ReAct harness, while LongHorizon-Harness retains its native Manage--Execute--Audit control structure.

\begin{itemize}
    \item \textbf{ALFWorld.}
    The agent receives a natural-language goal and an initial textual observation.
    At each step, it executes one command from the environment's admissible action set and receives the resulting observation.
    A case succeeds when the environment confirms that the goal conditions have been satisfied; otherwise, it terminates when the action budget is exhausted.

    \item \textbf{LOCA-Bench.}
    The agent receives a task instruction and a case-specific set of MCP tools for inspecting and modifying the task environment.
    Each tool call constitutes one environment action, and its result becomes the next observation.
    The agent invokes the \texttt{``claim\_done''} tool after completing the task, after which the official evaluator determines success; an uncompleted case terminates upon budget exhaustion.

    \item \textbf{RCA-100.}
    The agent initially receives the incident metadata, while metrics, logs, traces, events, alerts, and system topology must be acquired through diagnostic tools.
    Each tool call constitutes one environment action.
    The agent terminates the case by calling \texttt{``submit\_root\_cause''} with a fault type and a root-cause entity.
    We evaluate the two submitted fields separately and report joint accuracy when both are correct; cases without a valid submission before the action budget are treated as failures.

    \item \textbf{ClinDiag.}
    The agent initially observes only the patient's initial presentation.
    It acquires additional evidence by asking one specific question per action about medical history, physical examination, laboratory tests, radiographic tests, or other diagnostic procedures.
    The case terminates when the agent calls \texttt{``submit\_diagnosis''}, and the submitted diagnosis is evaluated against the reference diagnosis by the benchmark judge.
    Cases without a valid submission before the action budget are treated as failures.
\end{itemize}

\subsection{Metrics, Budgets, and Termination}
% 内容：
% 1. 定义各 benchmark 的评价指标。
% 2. 说明 ALFWorld 和 LOCA-Bench 的 Overall 是 micro average、
%    macro average，还是全部 cases 的总体成功率。
% 3. 定义 RCA-100 的 FT、RCL 和 Joint；特别说明 Joint 是否要求
%    FT 与 RCL 同时正确。
% 4. 定义 ClinDiag 的诊断正确性判定方式。
% 5. 报告每个 benchmark 的最大 environment-action budget。
% 6. 说明 success、failure、主动提交、无效动作、timeout 和达到
%    max steps 时的终止规则。
%
% 为什么要放：
% 主表结果和 cumulative-success 曲线都依赖这些定义。若分母、Joint
% 指标或终止规则不明确，不同方法之间的结果无法严格比较。
%
% 建议形式：
% 使用一张“Benchmark--Metric--Action Budget--Termination”配置表。
% 若使用 LLM-as-a-Judge 或人工评价，还应报告 judge prompt、重复评判方式
% 和冲突裁决规则。

Let \(y_i\in\{0,1\}\) indicate whether case \(i\) is successfully completed.
For a task category \(\mathcal{C}\), we compute its success rate as
\begin{equation}
    \operatorname{SR}(\mathcal{C})
    =
    \frac{1}{|\mathcal{C}|}
    \sum_{i\in\mathcal{C}} y_i.
    \label{eq:task-success-rate}
\end{equation}
The Overall results for ALFWorld and LOCA-Bench are micro-averaged over all evaluated cases rather than averaged across category-level success rates.
For LOCA-Bench, the main result pools cases across all environment lengths and random seeds, whereas the context-scaling analysis reports the same metric separately at each environment length.

For RCA-100, let \(y_i^{\mathrm{FT}}\) and \(y_i^{\mathrm{RCL}}\) indicate whether the submitted fault type and root-cause entity are correct, respectively.
We report
\begin{equation}
    \operatorname{FT}
    =
    \frac{1}{N}\sum_{i=1}^{N}y_i^{\mathrm{FT}},
    \qquad
    \operatorname{RCL}
    =
    \frac{1}{N}\sum_{i=1}^{N}y_i^{\mathrm{RCL}},
    \qquad
    \operatorname{Joint}
    =
    \frac{1}{N}\sum_{i=1}^{N}
    y_i^{\mathrm{FT}}y_i^{\mathrm{RCL}}.
    \label{eq:rca100-metrics}
\end{equation}
Thus, Joint counts a case as correct only when both the fault type and the root-cause entity are correctly identified.

For ClinDiag, diagnosis accuracy is the proportion of cases whose submitted final diagnosis is judged clinically equivalent to the reference diagnosis.
The judge accepts synonyms, standard abbreviations, and clinically equivalent formulations, while rejecting related but incorrect conditions or unsupported over-specific diagnoses.
One valid judge verdict is used for each case; malformed judge outputs are retried up to two times.

\begin{table*}[t]
\centering
\caption{Evaluation metrics, interaction budgets, and termination conditions.
All methods use the same budget within each benchmark.}
\label{tab:evaluation-protocols}
\small
\setlength{\tabcolsep}{4.2pt}
\renewcommand{\arraystretch}{1.15}
\begin{tabular}{@{}lp{4.0cm}p{3.0cm}p{4.3cm}@{}}
\toprule
\textbf{Benchmark}
& \textbf{Reported Metrics}
& \textbf{Maximum Budget}
& \textbf{Terminal Evaluation} \\
\midrule

ALFWorld
& Category and Overall task success rates
& 50 text actions
& The environment terminates the case when the goal is satisfied or another terminal state is reached. Success requires the environment's goal-completion signal. \\

LOCA-Bench
& Category and Overall task success rates
& 100 MCP tool calls
& Calling \texttt{claim\_done} triggers the official evaluator. A terminal score of at least \(1.0\) is counted as success. \\

RCA-100
& FT, RCL, and Joint accuracy
& 50 decision turns; at most 150 tool calls
& A valid \texttt{submit\_root\_cause} call terminates the case. FT and RCL are scored separately, and Joint requires both to be correct. \\

ClinDiag
& Diagnosis accuracy
& 30 tool calls
& A valid \texttt{submit\_diagnosis} call terminates the case, after which the submitted diagnosis is evaluated against the reference diagnosis. \\

\bottomrule
\end{tabular}
\end{table*}

A valid terminal submission ends the case regardless of whether its answer is correct.
Malformed submissions are rejected with an error observation and may be corrected within the remaining budget.
Other invalid actions or failed tool calls, including tool-level timeouts, likewise produce an error observation and consume the corresponding interaction opportunity without terminating the case.
\textbf{A case that reaches its maximum budget without successful completion is counted as a failure and remains in the denominator.}
Infrastructure failures outside the agent--environment interaction are retried and are not scored as task failures.

% =========================================================
% Appendix D: Implementation
% =========================================================
\section{Implementation and Reproducibility Details}
\label{app:implementation}
% 内容：集中说明模型、Agent 角色、baseline 实现、PoS 参数和重复运行设置。
% 作用：使正文能够保持凝练，同时回答公平性与可复现性方面的问题。

\subsection{Backbone and Agent Configuration}
\label{app:backbone-config}
% 内容：
% 1. 给出 Qwen3.7-Plus、Kimi-K3 和 GPT5.5 的准确模型名称、版本或 API snapshot。
% 2. 说明 Task Agent、Belief Manager 和 Belief Sentinel 分别使用哪个模型。
% 3. 说明这些角色是否始终使用与当前实验相同的 backbone。
% 4. 给出不同角色可调用的工具、可见上下文和执行顺序。
% 5. 说明是否启用模型原生 thinking，以及是否存在 API retry 或 fallback。
%
% 为什么要放：
% 如果 PoS 的不同模块使用了能力不同的模型，性能提升可能无法完全归因于
% framework 本身。角色配置必须透明。
%
% 建议形式：
% 使用一张“Role--Model--Input--Output--Tool Access”表。
% 注意 Sentinel 当前不再描述为 asynchronous，应与最终实现保持一致。

We evaluate all methods with three backbone models: \texttt{Qwen3.7-Plus}, \texttt{Kimi-K3}, and \texttt{GLM-5.3}.
The model is accessed through the DashScope OpenAI-compatible API (\texttt{https://dashscope.aliyuncs.com/compatible-mode/v1}); no fixed API snapshot is available.
For each backbone setting, all LLM-based components of \name and baselines use the same backbone as the corresponding baseline agents. 
We do not use a stronger auxiliary model for belief construction, validation, or trapping-aware recovery.
The Task Agent selects its action from the current task-conditioned belief state, which serves as its decision context.

The only model fixed independently of the evaluated backbone occurs in the medical diagnosis benchmark ClinDiag. 
Its provider and judge are treated as part of the benchmark environment rather than as components of the evaluated agent. 
To keep this environment identical across backbone settings, both are implemented with \texttt{Qwen3.7-plus} for all methods and backbones.
All agent-side model calls use deterministic decoding with temperature $0$. 
We set the maximum generation length to 512 tokens for ALFWorld, where each interaction primarily requires selecting an environment action, and to 4096 tokens for LOCA-Bench, RCA-100, and ClinDiag.

\subsection{Baseline Selection}
\label{app:baseline-selection}

We select baselines to cover representative forms of decision context while remaining applicable under a common cross-domain evaluation setting.
Raw Trajectory serves as the uncompressed reference and exposes the complete interaction history directly to the Agent.
Among memory-based approaches, ACON~\citep{kang2026acon}, PACE~\citep{wei2026pace}, and HiAgent~\citep{hu2025hiagent} represent complementary strategies based on history compression, adaptive context granularity, and hierarchical subgoal-oriented working memory, respectively.
We further include LongHorizon-Harness~\citep{ma2026longhorizonharness} as a closely related state-oriented baseline because it explicitly maintains an audited task state across long-horizon execution.
Together, these methods span raw history, compressed and selectively retained context, hierarchical memory, and explicit task-state maintenance, while remaining adaptable to the same four evaluation environments and backbone settings.

We do not attempt to exhaustively benchmark every method discussed in Section~\ref{sec:related-work}, as several approaches differ substantially in their training assumptions, control architecture, or task-specific state representation.
For example, SUPO~\citep{lu-etal-2026-beyond} jointly optimizes summarization and tool-use policies through reinforcement learning, while CaT~\citep{liu-etal-2026-context} and AgentFold~\citep{ye2026agentfold} rely on learned policies for actively manipulating the working context.
Several belief-oriented methods are likewise specialized to particular reasoning structures or task families.
In particular, Graph of States~\citep{luo2026graphstates} is designed for abductive reasoning and couples belief representation with a causal graph and a state machine that explicitly constrain hypothesis evolution, making its representation closely tied to diagnosis-style tasks rather than a common decision context that can be applied unchanged across all four benchmarks.
Similarly, Tru-POMDP~\citep{tang2025trupomdp} assumes structured hypotheses and belief-space planning for uncertain task planning.
We therefore prioritize baselines that provide representative and practically comparable alternatives to \name under a unified cross-domain evaluation protocol.

\subsection{Baseline Implementation}
\label{app:baseline-implementation}
% 内容：
% 1. 报告 Raw Trajectory、ACON、PACE、HiAgent、A-MEM 和其他最终采用的
%    baseline 的代码来源、版本或 commit。
% 2. 说明官方代码可以直接使用还是经过了跨 benchmark 适配。
% 3. 明确每个 baseline 提供给 Task Agent 的决策上下文。
% 4. 说明所有方法共享的 ReAct loop、工具接口和 environment-action budget。
% 5. 说明 baseline 中与原论文不同的实现选择及其原因。
%
% 为什么要放：
% Context-management 方法之间的区别主要体现在上下文构造方式。如果适配
% 过程不透明，审稿人难以判断比较是否公平。
%
% 建议形式：
% 使用一张“Method--Context Representation--Official Implementation--
% Adaptation”表，并在表后说明无法完全复现官方设置的特殊情况。

\paragraph{Shared evaluation protocol.}
We evaluate all baselines through the same benchmark adapters, tool
interfaces, task instructions, termination protocol, and environment-action
budgets described in Appendix~\ref{app:benchmarks}.
All LLM-based components follow the backbone configuration described in
Appendix~\ref{app:backbone-config}.
Because different policies naturally induce different interaction
trajectories, the methods do not receive identical observations; instead,
they share the same evidence-acquisition interface and differ in how the
evidence produced by their own interactions is represented and exposed for
subsequent decision making.
Raw Trajectory, ACON, PACE, and HiAgent are integrated into a common
ReAct harness~\citep{yao2023react}.
LongHorizon-Harness retains its native Manage--Execute--Audit control
structure, while using the same benchmark adapters, tools, backbone, and
environment-action budget.

\begin{table*}[t]
\centering
\caption{\textbf{Decision contexts and implementation provenance of the
evaluated baselines.}
All methods share the same benchmark interface and environment-action budget.
The main differences lie in how interaction evidence is represented and
exposed for subsequent decision making.}
\label{tab:baseline-implementation}

\small
\setlength{\tabcolsep}{3.2pt}
\renewcommand{\arraystretch}{1.12}

\begin{tabularx}{\textwidth}{
    @{}
    p{0.13\textwidth}
    p{0.29\textwidth}
    p{0.22\textwidth}
    X
    @{}
}
\toprule
\textbf{Method}
& \textbf{Decision Context}
& \textbf{Implementation Source}
& \textbf{Material Adaptation} \\
\midrule

Raw Trajectory
&
The task specification, current observation, and complete chronological
action--observation trajectory.
&
Shared ReAct implementation.
&
No compression, retrieval, or auxiliary context-management operation.
\\

ACON
&
A cumulative compressed summary together with the three most recent
interaction turns in their original form; before compression is triggered,
the full trajectory is retained.
&
Official implementation
(commit \texttt{d63f9ae18959})
~\citep{kang2026acon}.
&
Runtime compression is preserved, while failure-driven guideline
optimization and compressor distillation are omitted.
\\

PACE
&
A chronological sequence of historical chunks represented at full, detailed,
brief, or placeholder granularity according to predicted relevance, with the
two most recent chunks retained in full and selected chunks recoverable
through glimpse.
&
Reimplemented from the published method and Algorithm~1
~\citep{wei2026pace}.
&
The published repository was inaccessible during implementation;
\texttt{text-embedding-v4} replaces the original BGE-M3 encoder for relevance
scoring.
\\

HiAgent
&
Summaries of completed subgoals followed by the full action--observation
trajectory of the active subgoal; original trajectories of completed
subgoals remain retrievable when needed.
&
Official implementation
(commit \texttt{cebdd8e4eace})
~\citep{hu2025hiagent}.
&
The hierarchical working-memory mechanism is ported from AgentBoard to the
shared benchmark interface, with benchmark-level summary and folding
parameters.
\\

LongHorizon-Harness
&
An audited task state containing requirements, artifacts, and facts, together
with a bounded subtask contract for the current execution round.
&
Official implementation
~\citep{ma2026longhorizonharness}.
&
The native Manage--Execute--Audit architecture is preserved, with its
benchmark-facing tool interfaces adapted to our common evaluation setup.
\\

\bottomrule
\end{tabularx}
\end{table*}

\paragraph{Material implementation differences.}
The original baselines were developed for different environments and agent
protocols, so their end-to-end runners cannot be directly applied to all four
benchmarks.
We preserve the defining context-management mechanism of each method while
adapting its interaction layer to the common benchmark interfaces.
All methods receive the same task-independent background information, tool
documentation, and termination instructions.
Method-specific prompts contain only the instructions required to construct
or operate the corresponding context representation.
When an original response format is incompatible with the benchmark tool
interface, we provide format demonstrations containing only interface-level
examples and no evaluated cases or target answers.

\paragraph{ACON.}
ACON learns compression guidelines from contrasting successful and failed
trajectories and can further distill the optimized compressor
~\citep{kang2026acon}.
Our evaluation benchmarks do not provide dedicated training splits for
learning such benchmark-specific compression policies, while optimizing them
on evaluation cases would conflate inference-time context management with
task-specific training.
We therefore preserve ACON's runtime history-compression mechanism but use
fixed benchmark-level preservation guidelines without failure-driven
guideline optimization or compressor distillation.
The guidelines are case-independent and specify only generic categories of
information that should be retained.
Compression thresholds and output budgets are fixed at the benchmark level
and shared across all evaluated backbones.

\paragraph{PACE.}
PACE retains the interaction history in an external chunk store and exposes
each chunk at an adaptive granularity determined by its predicted relevance
to the next decision~\citep{wei2026pace}.
We preserve its four representation levels, relevance-based granularity
selection, chronological ordering, and glimpse mechanism.
Because the repository linked by the paper was inaccessible during our
implementation, we reproduce the method from its published equations,
default settings, and Algorithm~1.
The original BGE-M3 encoder is replaced with
\texttt{text-embedding-v4} for relevance scoring, while the evaluated
backbone generates the textual detailed and brief representations.

\paragraph{HiAgent.}
HiAgent organizes working memory around Agent-generated subgoals
~\citep{hu2025hiagent}.
The trajectory associated with the active subgoal is retained in full,
whereas completed subgoals are represented by summaries and their original
trajectories remain available for retrieval.
Since the official implementation is built on AgentBoard, we port the
subgoal-based working-memory mechanism into the shared benchmark interface
and convert its interaction protocol to the common structured action format.
Summary-length and folding parameters are adapted at the benchmark level and
then fixed across all cases and backbone models.

\paragraph{LongHorizon-Harness.}
LongHorizon-Harness differs from the other baselines in that it defines a complete Manage--Execute--Audit control architecture rather than only a context representation~\citep{ma2026longhorizonharness}.
At each round, the Manager constructs a bounded subtask from the audited task state, the Executor interacts with the environment to complete it, and the Auditor verifies the resulting state before verified requirements, artifacts, and facts are propagated to the next round.
We base our implementation on the official LongHorizon-Harness codebase and preserve its native Manager--Executor--Auditor workflow and audited task-state
representation.
We only adapt the benchmark-facing tool interfaces and action formats so that the method can operate through the same evaluation interfaces as the other baselines.
Any Auditor operation that queries the environment is counted against the same environment-action budget as an Executor action; purely internal Manager, Executor, or Auditor LLM calls are treated as computational overhead.

% =========================================================
% Appendix E: Supplementary experimental results and analyses
% =========================================================

\subsection{PoS Hyperparameters}
\label{app:pos-hyperparameters}

Table~\ref{tab:pos-hyperparameters} summarizes the hyperparameter configuration of \name.
Belief updates and local consistency audits are scheduled at every interaction step, while global audits are scheduled every 32 steps on LOCA-Bench and every 8 steps on the other benchmarks.
The diagnostic progress threshold applies to RCA-100 and ClinDiag, whereas execution progress is assessed by the Belief Sentinel.
Trapping detection begins once eight validated transitions are available, and recovery uses up to four recent transitions labeled as making no progress.

\begin{table}[ht]
\centering
\caption{Hyperparameter configuration of \name. Intervals are measured in interaction steps.}
\label{tab:pos-hyperparameters}
\small
\setlength{\tabcolsep}{6pt}
\renewcommand{\arraystretch}{1.1}
\begin{tabular}{@{}lc@{}}
\toprule
\textbf{Parameter} & \textbf{Value} \\
\midrule
Belief update interval & 1 \\
Local audit interval & 1 \\
Global audit interval (LOCA-Bench) & 32 \\
Global audit interval (other benchmarks) & 8 \\
Recent transitions supplied for local auditing & 4 \\
\midrule
Detection window size \(K\) & 8 \\
Maximum recurrence lag \(L\) & 4 \\
Recurrence distance threshold \(\epsilon_W\) & 0.15 \\
Diagnostic progress threshold \(\epsilon\) & 0.30 \\
Health threshold \(\theta_H\) & 0.25 \\
Cycle recurrence threshold \(\theta_R\) & 0.75 \\
Maximum recent non-progress transitions for recovery & 4 \\
\bottomrule
\end{tabular}
\end{table}

\section{Fine-Grained Main Results}
\label{app:fine-grained-results}

\begin{table*}[t]
\centering
\caption{Main results across four benchmarks (\%).
We report task success for ALFWorld and LOCA-Bench, failure triage (FT), root-cause localization (RCL), and joint accuracy for RCA-100, and diagnosis accuracy for ClinDiag.
Env. Ops denotes Environment Operations.}
\label{tab:fine-grained-main-results}

\vspace{2mm}

\footnotesize
\setlength{\tabcolsep}{1.45pt}
\renewcommand{\arraystretch}{1.12}

\renewcommand{\BackboneHeader}[2]{%
  \rowcolor{modelgray}%
  \multicolumn{15}{c}{%
    \textbf{Backbone: }\modelicon{#1}\textbf{#2}%
  }\\[0.10ex]%
  \specialrule{0.45pt}{0pt}{0.35ex}%
}

\resizebox{\textwidth}{!}{%
\begin{tabular}{@{}l*{14}{c}@{}}
\toprule

\multirow{2}{*}{\textbf{Method}} &
\multicolumn{4}{c}{\textbf{ALFWorld}} &
\multicolumn{4}{c}{\textbf{LOCA-Bench}} &
\multicolumn{3}{c}{\textbf{RCA-100}} &
\multicolumn{3}{c}{\textbf{ClinDiag}} \\

\cmidrule(lr){2-5}
\cmidrule(lr){6-9}
\cmidrule(lr){10-12}
\cmidrule(lr){13-15}

& Place & Transform & Inspect & Overall
& Cross-source & Analytics & Env. Ops & Overall
& FT & RCL & Joint
& Common & Rare & Overall \\

\specialrule{0.7pt}{0pt}{0.25ex}

\BackboneHeader{qwen}{Qwen3.7-Plus}

Raw Trajectory
& 92.68 & 42.67 & 77.78 & 62.69
& 39.43 & 65.71 & 25.71 & 43.62
& 26.21 & 68.93 & 24.27
& 37.75 & 40.07 & 38.91 \\

ACON
& \textbf{97.56} & 48.00 & 72.22 & 66.42
& 49.14 & \textbf{75.43} & 25.14 & 49.90
& 30.10 & 68.93 & 27.18
& 39.07 & 40.40 & 39.74 \\

PACE
& 92.68 & 52.00 & 77.78 & 67.91
& 4.57 & 40.57 & 6.86 & 17.33
& 33.01 & 66.99 & 28.16
& 43.04 & 40.73 & 41.89 \\

HiAgent
& 90.24 & 48.00 & 88.89 & 66.42
& 18.86 & 40.00 & 14.86 & 24.57
& 32.04 & \textbf{76.70} & 28.16
& 39.74 & 40.40 & 40.07 \\

LongHorizon-Harness
& 95.12 & 56.00 & 88.89 & 72.39
& 53.71 & 72.57 & 31.43 & 52.57
& 31.07 & 75.73 & 26.21
& 41.72 & 39.40 & 40.56 \\

\specialrule{0.35pt}{0.35ex}{0.20ex}

\ColorFirstCell{posmain}{\PoSMethod}%
& \ColorMiddleCell{posmain}{\textbf{97.56}}
& \ColorMiddleCell{posmain}{\textbf{81.33}}
& \ColorMiddleCell{posmain}{\textbf{100.00}}
& \ColorMiddleCell{posmain}{\textbf{88.81}}
& \ColorMiddleCell{posmain}{\textbf{58.86}}
& \ColorMiddleCell{posmain}{74.29}
& \ColorMiddleCell{posmain}{\textbf{36.00}}
& \ColorMiddleCell{posmain}{\textbf{56.38}}
& \ColorMiddleCell{posmain}{\textbf{39.81}}
& \ColorMiddleCell{posmain}{67.96}
& \ColorMiddleCell{posmain}{\textbf{38.83}}
& \ColorMiddleCell{posmain}{\textbf{48.01}}
& \ColorMiddleCell{posmain}{\textbf{42.05}}
& \ColorLastCell{posmain}{\textbf{45.03}} \\

\ColorFirstCell{posablation}{%
  \hspace{0.35em}\quad - Consistency Validation%
}%
& \ColorMiddleCell{posablation}{92.68}
& \ColorMiddleCell{posablation}{65.33}
& \ColorMiddleCell{posablation}{66.67}
& \ColorMiddleCell{posablation}{73.88}
& \ColorMiddleCell{posablation}{42.29}
& \ColorMiddleCell{posablation}{69.14}
& \ColorMiddleCell{posablation}{22.29}
& \ColorMiddleCell{posablation}{44.57}
& \ColorMiddleCell{posablation}{33.01}
& \ColorMiddleCell{posablation}{66.99}
& \ColorMiddleCell{posablation}{31.07}
& \ColorMiddleCell{posablation}{47.35}
& \ColorMiddleCell{posablation}{41.72}
& \ColorLastCell{posablation}{44.54} \\

\ColorFirstCell{posablation}{%
  \hspace{0.35em}\quad - Trapping Diagnosis%
}%
& \ColorMiddleCell{posablation}{85.37}
& \ColorMiddleCell{posablation}{70.67}
& \ColorMiddleCell{posablation}{55.56}
& \ColorMiddleCell{posablation}{73.13}
& \ColorMiddleCell{posablation}{48.00}
& \ColorMiddleCell{posablation}{71.43}
& \ColorMiddleCell{posablation}{25.14}
& \ColorMiddleCell{posablation}{48.19}
& \ColorMiddleCell{posablation}{35.92}
& \ColorMiddleCell{posablation}{63.11}
& \ColorMiddleCell{posablation}{33.98}
& \ColorMiddleCell{posablation}{45.03}
& \ColorMiddleCell{posablation}{39.74}
& \ColorLastCell{posablation}{42.38} \\

\specialrule{0.6pt}{0.40ex}{0.25ex}

\BackboneHeader{kimi}{Kimi-K3}

Raw Trajectory
& 97.56 & 68.00 & \textbf{100.00} & 81.34
& 88.00 & 84.57 & 36.57 & 69.71
& 38.83 & 72.82 & 36.89
& 50.66 & 47.68 & 49.17 \\

ACON
& \textbf{100.00} & 66.67 & \textbf{100.00} & 81.34
& 90.29 & 86.86 & 35.43 & 70.86
& 41.75 & \textbf{79.61} & 40.78
& 39.07 & 38.08 & 38.58 \\

PACE
& \textbf{100.00} & 76.00 & 83.33 & 84.33
& 52.00 & 50.86 & 21.71 & 41.52
& 36.89 & 74.76 & 35.92
& 50.66 & 49.67 & 50.17 \\

HiAgent
& 97.56 & 81.33 & \textbf{100.00} & 88.81
& 63.43 & 59.43 & 25.71 & 49.52
& 42.72 & 78.64 & 41.75
& 50.99 & 46.03 & 48.51 \\

LongHorizon-Harness
& \textbf{100.00} & 86.67 & 94.44 & 91.79
& 84.57 & 80.00 & 38.29 & 67.62
& 43.69 & 76.70 & 38.83
& 51.32 & \textbf{52.32} & 51.82 \\

\specialrule{0.35pt}{0.35ex}{0.20ex}

\ColorFirstCell{posmain}{\PoSMethod}%
& \ColorMiddleCell{posmain}{\textbf{100.00}}
& \ColorMiddleCell{posmain}{\textbf{89.33}}
& \ColorMiddleCell{posmain}{\textbf{100.00}}
& \ColorMiddleCell{posmain}{\textbf{94.03}}
& \ColorMiddleCell{posmain}{\textbf{92.57}}
& \ColorMiddleCell{posmain}{\textbf{89.14}}
& \ColorMiddleCell{posmain}{\textbf{41.14}}
& \ColorMiddleCell{posmain}{\textbf{74.29}}
& \ColorMiddleCell{posmain}{\textbf{51.46}}
& \ColorMiddleCell{posmain}{76.70}
& \ColorMiddleCell{posmain}{\textbf{51.46}}
& \ColorMiddleCell{posmain}{\textbf{57.28}}
& \ColorMiddleCell{posmain}{\textbf{52.32}}
& \ColorLastCell{posmain}{\textbf{54.80}} \\

\ColorFirstCell{posablation}{%
  \hspace{0.35em}\quad - Consistency Validation%
}%
& \ColorMiddleCell{posablation}{97.56}
& \ColorMiddleCell{posablation}{88.00}
& \ColorMiddleCell{posablation}{94.44}
& \ColorMiddleCell{posablation}{91.79}
& \ColorMiddleCell{posablation}{84.57}
& \ColorMiddleCell{posablation}{81.71}
& \ColorMiddleCell{posablation}{36.00}
& \ColorMiddleCell{posablation}{67.43}
& \ColorMiddleCell{posablation}{48.54}
& \ColorMiddleCell{posablation}{75.73}
& \ColorMiddleCell{posablation}{48.54}
& \ColorMiddleCell{posablation}{56.95}
& \ColorMiddleCell{posablation}{51.99}
& \ColorLastCell{posablation}{54.47} \\

\ColorFirstCell{posablation}{%
  \hspace{0.35em}\quad - Trapping Diagnosis%
}%
& \ColorMiddleCell{posablation}{97.56}
& \ColorMiddleCell{posablation}{82.67}
& \ColorMiddleCell{posablation}{94.44}
& \ColorMiddleCell{posablation}{88.81}
& \ColorMiddleCell{posablation}{88.00}
& \ColorMiddleCell{posablation}{85.14}
& \ColorMiddleCell{posablation}{35.43}
& \ColorMiddleCell{posablation}{69.52}
& \ColorMiddleCell{posablation}{46.60}
& \ColorMiddleCell{posablation}{72.82}
& \ColorMiddleCell{posablation}{46.60}
& \ColorMiddleCell{posablation}{52.98}
& \ColorMiddleCell{posablation}{48.68}
& \ColorLastCell{posablation}{50.83} \\

\specialrule{0.6pt}{0.40ex}{0.25ex}

\BackboneHeader{zhipu}{GLM-5.3}

Raw Trajectory
& 97.56 & 89.33 & \textbf{100.00} & 93.28
& 80.00 & 76.00 & 32.57 & 62.86
& 34.95 & 71.84 & 32.04
& 50.00 & 43.71 & 46.85 \\

ACON
& 97.56 & 92.00 & \textbf{100.00} & 94.78
& 85.14 & 81.14 & 31.43 & 65.90
& 33.98 & 77.67 & 33.98
& 49.01 & 43.38 & 46.19 \\

PACE
& 97.56 & 92.00 & \textbf{100.00} & 94.78
& 48.00 & 47.43 & 19.43 & 38.29
& 39.81 & \textbf{78.64} & 37.86
& 47.68 & 39.07 & 43.38 \\

HiAgent
& \textbf{100.00} & 90.67 & \textbf{100.00} & 94.78
& 59.43 & 54.86 & 23.43 & 45.90
& 43.69 & 74.76 & 40.78
& 49.34 & 40.40 & 44.87 \\

LongHorizon-Harness
& 97.56 & 93.33 & \textbf{100.00} & 95.52
& 81.71 & 78.29 & 33.14 & 64.38
& 38.83 & 73.79 & 31.07
& 48.68 & 42.38 & 45.53 \\

\specialrule{0.35pt}{0.35ex}{0.20ex}

\ColorFirstCell{posmain}{\PoSMethod}%
& \ColorMiddleCell{posmain}{\textbf{100.00}}
& \ColorMiddleCell{posmain}{\textbf{94.67}}
& \ColorMiddleCell{posmain}{\textbf{100.00}}
& \ColorMiddleCell{posmain}{\textbf{97.01}}
& \ColorMiddleCell{posmain}{\textbf{88.00}}
& \ColorMiddleCell{posmain}{\textbf{85.71}}
& \ColorMiddleCell{posmain}{\textbf{38.86}}
& \ColorMiddleCell{posmain}{\textbf{70.86}}
& \ColorMiddleCell{posmain}{\textbf{50.49}}
& \ColorMiddleCell{posmain}{75.73}
& \ColorMiddleCell{posmain}{\textbf{49.51}}
& \ColorMiddleCell{posmain}{\textbf{54.97}}
& \ColorMiddleCell{posmain}{\textbf{49.34}}
& \ColorLastCell{posmain}{\textbf{52.15}} \\

\ColorFirstCell{posablation}{%
  \hspace{0.35em}\quad - Consistency Validation%
}%
& \ColorMiddleCell{posablation}{97.56}
& \ColorMiddleCell{posablation}{93.33}
& \ColorMiddleCell{posablation}{94.44}
& \ColorMiddleCell{posablation}{94.78}
& \ColorMiddleCell{posablation}{79.43}
& \ColorMiddleCell{posablation}{75.43}
& \ColorMiddleCell{posablation}{33.14}
& \ColorMiddleCell{posablation}{62.67}
& \ColorMiddleCell{posablation}{49.51}
& \ColorMiddleCell{posablation}{74.76}
& \ColorMiddleCell{posablation}{47.57}
& \ColorMiddleCell{posablation}{53.97}
& \ColorMiddleCell{posablation}{49.01}
& \ColorLastCell{posablation}{51.49} \\

\ColorFirstCell{posablation}{%
  \hspace{0.35em}\quad - Trapping Diagnosis%
}%
& \ColorMiddleCell{posablation}{97.56}
& \ColorMiddleCell{posablation}{89.33}
& \ColorMiddleCell{posablation}{94.44}
& \ColorMiddleCell{posablation}{92.54}
& \ColorMiddleCell{posablation}{85.14}
& \ColorMiddleCell{posablation}{79.43}
& \ColorMiddleCell{posablation}{32.57}
& \ColorMiddleCell{posablation}{65.71}
& \ColorMiddleCell{posablation}{45.63}
& \ColorMiddleCell{posablation}{71.84}
& \ColorMiddleCell{posablation}{42.72}
& \ColorMiddleCell{posablation}{50.99}
& \ColorMiddleCell{posablation}{46.03}
& \ColorLastCell{posablation}{48.51} \\

\bottomrule
\end{tabular}%
}
\end{table*}

Table~\ref{tab:fine-grained-main-results} expands the aggregate results in Table~\ref{tab:main-results} into task categories and diagnostic dimensions.
This breakdown reveals which task requirements benefit from explicit belief maintenance, how validation and recovery contribute to different aspects of performance, and where challenges remain.

\paragraph{(1) Explicit belief particularly benefits state tracking and diagnostic disambiguation.}
A key benefit of explicit belief is that observations from different steps jointly inform an updated understanding of the current world.
By maintaining this understanding together with unresolved gaps, \name connects what the agent has already established to what its next action must accomplish.
This is particularly useful when successful execution depends on preserving state changes across actions, or when diagnosis requires combining several signals to distinguish competing hypotheses.
The concentration of ALFWorld gains in \texttt{Transform} supports this interpretation, with \texttt{Qwen3.7-Plus} success increasing from \(56.00\%\) to \(81.33\%\), while \texttt{Place} matches the strongest baseline at \(97.56\%\).
Our RCA-100 trace analysis further shows how belief integrates ``request patterns'' and ``resource-utilization'' signals around an identified entity, allowing the agent to rule out failure types that conflict with the accumulated observations.
Most failure-type corrections occur in cases where Raw Trajectory already identifies the correct entity but misclassifies its failure, including 14 cases with \texttt{Kimi-K3} and 16 with \texttt{GLM-5.3}.
Together with the trace analysis, this highlights how explicit belief consolidates failure characteristics around an entity, while the Active Gap directs further investigation toward evidence that distinguishes competing hypotheses.

\paragraph{(2) Validation and recovery support different aspects of diagnostic performance.}
On RCA-100, removing Consistency Validation reduces RCL by only \(0.97\) points under each backbone, but lowers FT by \(6.80\) points with Qwen3.7-Plus.
This contrast highlights the importance of validating inferred fault states even when the agent can still locate the incident.
Removing Trapping Diagnosis and its associated recovery has a larger effect on localization, reducing RCL by \(3.88\)--\(4.85\) points.
The distinction also appears in ClinDiag's \texttt{Rare} subset, where removing validation lowers accuracy by \(0.33\) points under each backbone, whereas removing recovery causes losses of \(2.31\)--\(3.64\) points.
These results show that the two mechanisms address different needs within diagnosis.
Validation checks whether belief updates remain consistent and supported, while recovery redirects investigation when the current course of interaction no longer resolves diagnostic uncertainty.

\paragraph{(3) Better decision context does not replace execution skills or domain knowledge.}
Explicit belief construction and maintenance improve how task information is represented and updated, but successful completion also depends on the model's ability to act on that information.
Although \name achieves the highest success on LOCA-Bench's \texttt{Env.Ops} category under every backbone, performance remains between \(36.00\%\) and \(41.14\%\).
These tasks also demand reliable tool use and action sequencing, as identifying the required state change is different from executing the operations that produce it.
Our ClinDiag trace analysis reveals a related limitation, with many incorrect diagnoses arising after the agent receives correct information but lacks the medical knowledge needed to interpret it.
Even an accurate representation of the current World State is insufficient when the agent lacks the execution skills or domain knowledge required to complete the task.
Addressing these limitations requires improvements in tool-use competence and domain-specific reasoning alongside belief maintenance, rather than relying on context management alone.

% =========================================================
% Appendix F: Case study and failure analysis
% =========================================================
\section{Case Study}
\label{app:case-study}
% 内容：使用一个真实案例展示 Belief、Active Gap、Sentinel、Belief Health
% 和 Recovery 如何在实际交互中共同工作，并补充方法的主要失败模式。
% 作用：案例用于解释机制，而跨 benchmark 的普适性由定量实验支撑。
% 建议展开正文 teaser 中的真实 ALFWorld hot-mug case，使用一张关键
% transition 表展示必要的 action、observation、Belief update、progress
% signal、trapping diagnosis 和 recovery constraint；无需打印完整轨迹。
% 不再分别设置四个 benchmark 的 case-study 小节。

% 沿用已有的 posStudyBlock 环境和颜色定义。
% 不使用 figure、caption 或 minipage，内容随正文排布。

We use RCA-100 case \texttt{t024} with \texttt{Kimi-K3} to illustrate \name's trapping diagnosis and recovery.
The agent observes abnormal inventory-service behavior but remains unable to distinguish database waiting from JVM-side processing despite continued investigation.
\name identifies ``Static Stagnation'' with a persistent epistemic gap and combines constraints against ineffective repetition with a requirement to seek evidence that distinguishes these explanations.
The agent subsequently obtains CPU observations and integrates them with the earlier GC evidence, revising its belief and ultimately submitting the correct diagnosis.
The case below traces this process from the unresolved diagnostic question to targeted recovery and belief revision.

\par
\begingroup
\small
\raggedright

\noindent
\textbf{RCA-100: recovering from stalled diagnosis}
\hfill
\textbf{Case \texttt{t024} | Kimi-K3}

\medskip
The agent must identify the fault entity and failure type.
This excerpt covers one of three recovery episodes in a 28-action run.
Step indices are zero-based.

\begin{posStudyBlock}{posCaseBlue}
  {1. Investigation continues, but the diagnostic question remains unresolved}

\textbf{Current belief.}
Slow inventory requests coincide with elevated garbage-collection
(GC) activity.

\textbf{Active epistemic gap.}
Are the delays caused by database waiting or processing inside
the inventory JVM?
Further log searches do not distinguish these explanations.

\textbf{Trapping diagnosis.}
At step \(22\), the same gap persists with little change in the belief.
PoS identifies \emph{Belief Trapping} pattern as
\textcolor{posCaseOrange}{\textbf{Static Stagnation + Epistemic Gap}}.

\end{posStudyBlock}

\begin{posStudyBlock}{posCaseOrange}
  {2. Factorized recovery redirects investigation toward the unresolved gap}

\textbf{Pattern constraint.}
Avoid repeating the ineffective action under the unchanged belief.

\textbf{Gap constraint.}
Seek evidence that distinguishes database waiting from
processing inside the JVM.

\textbf{Agent-selected investigation.}
The agent chooses to inspect CPU activity.
The recovery constraints specify what the investigation must resolve,
rather than prescribe a tool call or provide the diagnosis.

\end{posStudyBlock}

\begin{posStudyBlock}{posCaseGreen}
  {3. New evidence changes the belief and advances diagnosis}

\textbf{New evidence.}
Recovery initially produces three further low-progress actions.
Subsequent CPU queries reveal increased activity during the incident.

\textbf{Belief revision.}
The agent combines the CPU evidence with the earlier GC observations,
increases its support for JVM-side processing, and reduces its support
for database waiting.
The targeted epistemic gap is recorded as resolved at step \(26\).

\textbf{Outcome.}
At step \(27\), the agent submits
\textcolor{posCaseGreen}{%
  \textbf{\texttt{inventory / memoryPressure}}},
matching both the entity and failure type in the benchmark label.

\end{posStudyBlock}

\textit{Scope.}
The case illustrates how targeted recovery restores diagnostic progress.
CPU evidence supports the recorded belief revision but does not uniquely
determine the final failure type.

\par
\endgroup

\section{Limitations and Future Directions}

Two limitations motivate future work on the efficiency of belief maintenance and the reliability of knowledge-dependent reasoning in PoS.

\paragraph{(1) Cost of explicit belief maintenance.}
The performance gains of \name come with additional token and computational costs. 
Its natural language form structured Belief States require repeated text processing and generation during maintenance and subsequent reasoning. 
Repeatedly expressing and revising beliefs in language introduces redundancy and may cause information omission or semantic distortion. 
Future work could explore task-conditioned \textbf{latent world-state representations} that encode decision-relevant information more compactly and support incremental updates without repeatedly expanding the state into text. 
The key is to encode information about the current world state in a latent representation with high information density that can be efficiently propagated and updated within the model.
We aim to preserve an accurate understanding of the current world state while substantially reducing the token and computational costs of belief maintenance.

\paragraph{(2) Dependence on domain and task knowledge.}
Although \name receives external observations through tools, it does not systematically integrate external domain-knowledge retrieval into belief construction and reasoning. 
Its ability to identify relevant evidence and interpret observations therefore depends largely on the backbone model's existing knowledge. 
Insufficient knowledge may lead to omitted facts or incorrect causal and diagnostic judgments, even when the resulting Belief State is internally consistent. 
Future work could incorporate \textbf{domain-knowledge retrieval} into belief updates and evidence interpretation, helping the agent determine what information to retain and how it supports or contradicts competing explanations. 
By incorporating relevant domain knowledge, we aim to improve the accuracy of Belief States and the reliability of subsequent reasoning in specialized tasks.